\documentclass[preprint,12pt]{elsarticle}

\usepackage{amsmath}
\usepackage{amssymb}
\usepackage{amsfonts}
\usepackage{bm}
\usepackage{graphicx}
\usepackage{booktabs} 
\usepackage{multirow}
\usepackage{tabularx}
\usepackage{adjustbox}
\usepackage{array}
\usepackage{subcaption}
\usepackage{textcomp}
\usepackage{xcolor}
\usepackage{url}
\usepackage{hyperref}

\usepackage{algorithm}
\usepackage[noend]{algpseudocode}

\journal{Information Fusion}

\begin{document}

\begin{frontmatter}

\title{Missingness-Conditioned Spatio-Temporal Fusion for Human Activity Recognition from Low-Quality Wearable Sensor Data}


\author[durham]{Jiangtao Fan}
\ead{jiangtao.fan@durham.ac.uk}

\author[durham]{Anish Jindal\corref{cor1}}
\ead{anish.jindal@durham.ac.uk}

\author[durham]{Amir Atapour-Abarghouei}
\ead{amir.atapour-abarghouei@durham.ac.uk}

\cortext[cor1]{Corresponding author.}

\affiliation[durham]{organization={Department of Computer Science, Durham University},
            city={Durham},
            postcode={DH1 3LE},
            country={United Kingdom}}

\begin{abstract}
Body-worn sensor networks capture human activities through complementary views, where nodes, modalities, and axes partially describe the underlying motion. In real deployments, hardware faults, wireless dropouts, and loose attachments often cause entire channels or nodes to go missing, while environmental interference degrades the remaining signals. The common impute-then-recognize pipeline introduces reconstruction errors that cannot be observed or corrected at inference. We propose the Missingness-Conditioned Spatio-Temporal Network (MCSTN), an end-to-end framework for robust activity recognition under imperfect sensing. MCSTN has three components: a structured imperfection generator that simulates point, temporal-block, axis, modality, and node failures, plus noise at varying signal-to-noise levels; a mask-aware representation that conditions node embeddings, attention, and temporal aggregation on the availability mask; and a complete-to-imperfect multi-view learning scheme where an exponential-moving-average teacher guides multiple degraded student views through semantic consistency. The encoder models temporal and cross-feature dependencies in parallel: a bi-directional selective state-space branch captures long-range temporal dependencies, a transposed-attention branch models cross-channel correlations, and adaptive weights fuse their outputs. At inference, the generator, teacher, and consistency loss are removed, so MCSTN predicts directly from the observed window and its mask without reconstruction or noise-level estimation. On PAMAP2, Opportunity, and WISDM, MCSTN achieves average macro-F1 scores of 94.24\%, 91.67\%, and 96.22\% under imperfect observations, the best among all compared methods on each dataset, and the best in every failure mode on Opportunity. It requires fewer than one million parameters, with 3.78~ms inference latency per sample.
\end{abstract}


\end{frontmatter}

\section{Introduction}
\label{sec:introduction}

Sensor-based Human Activity Recognition (HAR) aims to identify human activities
from body-worn inertial and physiological sensors. It supports a wide range of
applications, including healthcare monitoring, rehabilitation, elderly care,
and context-aware services in Internet-of-Things (IoT) environments\cite{li_efficient_2024,
sun_actiformer_2025}. Sensor-based HAR is naturally a multi-view sensing problem. One activity can be observed by sensors placed at different body locations, by different sensing modalities, and by multiple measurement axes within each modality. These observations provide complementary information. For example, wrist sensors can capture hand
movements, ankle sensors measure gait dynamics, and chest sensors can track body posture. Consequently, accurate activity recognition relies on effectively fusing these complementary views rather than processing isolated data streams \cite{qiu_multi-sensor_2022, islam_multi-level_2023, li_human_2023}. 

Recent deep learning methods have reduced the dependence on handcrafted statistical and frequency-domain features, which often require expert knowledge and may generalise poorly across subjects and sensor placements \cite{kumar_human_2024}. Convolutional networks \cite{ignatov_real-time_2018}, recurrent networks \cite{zeng_understanding_2018}, and attention-based models \cite{haresamudram_masked_2020} can learn  hierarchical spatio-temporal representations directly from sensor signals and have achieved strong performance on standard HAR benchmarks. However, most existing methods assume ideal sensing conditions, where all channels are fully observed across all time steps. This assumption is difficult to
maintain in real-world deployment \cite{yang_multiscale_2024, liu_disentangling_2024}. Sensors may become loose, batteries may run out, wireless transmission may fail, and individual sensing units may become noisy or unavailable. Low-quality sensor data may result from both missing measurements and corrupted measurements. These two forms should be treated separately. Missingness determines which measurements are available. It is often structured by the sensor configuration rather than occurring independently. For example, battery failure may remove an entire sensor node, an axis fault may affect one channel over a long period, and a failed gyroscope may remove one modality while other sensors at the same location remain available. By contrast, environmental degrades the quality of the measurements that exist. These different failure patterns have different effects on recognition. For example, randomly removing 30\% of individual samples does not represent the same information loss with removing 30\% of the input through a complete sensor node.

The common approach to handling low-quality sensor data relies on a two-stage strategy: imputation followed by classification. Existing methods employ statistical interpolation, autoencoders \cite{goubeaud_using_2021}, generative adversarial networks \cite{hussein_sensorgan_2024, guo_data_2019}, or recurrent/attention-based imputers to reconstruct missing entries before feeding the completed data into a downstream classifier \cite{pires_improving_2020, sun_ds-hbi_2025}. Although this strategy maintains compatibility with standard classifiers, it also introduces an additional source of error. During inference, imputation errors are neither observable nor bounded. When corruption patterns deviate from training assumptions, the imputer generates inaccurate values, propagating errors to the classifier \cite{liu_modality_2024}. Conversely, corruption-robust representation learning methods \cite{cheng_maskcae_2024, liu_disentangling_2024, hossain_method_2020} bypass data reconstruction, but they often focus only on random point-wise noise and ignore group-level missingness patterns.

We treat the availability pattern as an input to fusion rather than a defect to be repaired. Three design choices follow from this view. Training degradation should follow the physical structure of the sensor network, so failures at the point, temporal-block, axis, modality, and node levels are generated explicitly instead of being approximated by random dropout. The recognition model should see the availability mask, because after normalisation a value of zero can be a valid measurement. Degraded views should be anchored to the most complete available observation of the same activity window, since aligning two independently degraded views only asks two incomplete representations to agree.

Based on these choices, we propose the Missingness-Conditioned Spatio-Temporal Network (MCSTN), shown in Fig.~\ref{fig:overview}. During training, a structured imperfection generator creates several imperfect views of each window by combining point, temporal-block, axis, modality, and node failures at different intensities, and adds noise on a separate signal-to-noise schedule. Each view is encoded together with its availability mask by a parallel spatio-temporal encoder: a bi-directional selective state-space branch models temporal dependencies, a mask-aware transposed-attention branch models interactions between feature dimensions, and adaptive weights fuse the two. An exponential-moving-average (EMA) teacher processes the reference view and guides the student views through a semantic consistency objective. The generator and the teacher exist only during training, so at inference MCSTN classifies the observed window and its mask directly.

Our contributions are as follows.

\begin{itemize}

\item We formulate robust HAR as a missingness-conditioned multi-view fusion problem instead of a two-stage imputation-classification pipeline, so that no reconstructed values enter the classifier at inference.

\item We introduce a structured imperfection generator that models five types of sensor failure: point, temporal block, axis, modality, and node. Missingness and measurement noise are generated separately. Used as training augmentation, the generator also improves every competing model evaluated on Opportunity.

\item We design a mask-aware parallel encoder that models temporal dynamics and
cross-feature dependencies in separate branches. The availability mask is used
in the input representation, attention computation, and temporal aggregation,
and the two branches are combined through input-dependent fusion.

\item We propose a complete-to-imperfect multi-view learning scheme in which an EMA teacher processes the reference view and supervises the imperfect student views. On PAMAP2, Opportunity, and WISDM, MCSTN obtains the highest average macro-F1 under structured sensor failures at a computational cost comparable to the evaluated baselines.

\end{itemize}

The remainder of this paper is organised as follows.
Section~\ref{sec:relatedworks} reviews deep models for sensor-based HAR,
learning under incomplete and noisy sensor streams, and consistency-based
multi-view representation learning. Section~\ref{sec:method} presents MCSTN.
Section~\ref{sec:experiments} describes the experimental protocol and reports
the results, and Section~\ref{conclusion} concludes the paper.

\section{Related Works}
\label{sec:relatedworks}

\subsection{Deep Learning for Sensor-based HAR}

Deep learning has become the main approach to sensor-based HAR. Convolutional networks (CNNs) are widely used to extract local temporal patterns and correlations across sensor channels. Yu \textit{et al.} \cite{yu_human_2024} combine convolutions with different receptive fields to capture motion patterns at multiple temporal scales. Recurrent networks (RNNs) focus more directly on temporal dependencies. Zeng \textit{et al.} \cite{zeng_understanding_2018} combine multiple LSTM networks to model complementary temporal patterns, while Uddin \textit{et al.} \cite{uddin_body_2020} use deep recurrent networks to fuse body-sensor signals for behaviour recognition in healthcare applications. Hybrid architectures such as DeepConvLSTM \cite{ordonez_deep_2016, singh_deep_2020} combine convolutional feature extraction with recurrent sequence modelling and remain widely used baselines in HAR.

Attention-based models have also been introduced to capture long-range dependencies in sensor sequences. Teng \textit{et al.} \cite{teng_innovative_2024} combine spatial and temporal attention for spatio-temporal feature learning. Masked Transformer pre-training has been applied to inertial signals \cite{haresamudram_masked_2020}, while two-stream
Transformers model complementary sensor representations in separate branches \cite{xiao_two-stream_2022}. Pang \textit{et al.} \cite{pang_cross-attention_2024} further combine cross-attention with multi-scale convolution to capture temporal patterns at different scales. Standard self-attention, however, scales quadratically with sequence length, which raises the cost of on-device inference.

Selective state-space models (SSMs) \cite{gu_mamba_2023, miao_comprehensive_2025} model long sequences in linear time, and HARMamba \cite{li_harmamba_2024} applies them to wearable HAR. Unlike single-stream architectures that couple spatial and temporal feature extraction, MCSTN separates the two: a bi-directional selective state-space branch models temporal dynamics in linear time, and a transposed-attention branch models interactions between feature dimensions with an attention map whose size does not depend on sequence length. The availability mask enters the shared input embedding, and the attention branch and the temporal aggregation also use it directly.

\subsection{Learning from Incomplete and Noisy Sensor Streams}

Existing methods for incomplete sensor data can be broadly divided into reconstruction-based approaches and representation-based approaches. Reconstruction-based methods estimate missing measurements before or during the recognition process. Common techniques include statistical interpolation, autoencoders \cite{goubeaud_using_2021}, generative adversarial networks \cite{hussein_sensorgan_2024, guo_data_2019}, and models that jointly optimise imputation and prediction \cite{sun_ds-hbi_2025}.

Missing-data modelling has also been studied extensively for multivariate time
series. GRU-D \cite{che_recurrent_2018} incorporates missingness indicators and time intervals directly into recurrent modelling. BRITS \cite{cao_brits_2018} estimates missing values with bidirectional recurrent dynamics, while SAITS \cite{du_saits_2023} uses self-attention for multivariate time-series imputation. More recent graph-based methods model dependencies across variables and time when estimating missing observations \cite{zheng_graph_2024}. These methods show that both temporal and cross-variable structure are important when handling incomplete sequences. However, methods that rely on reconstructed values can introduce additional uncertainty before classification, particularly when the test-time missingness pattern differs from that used during training. Similar sensitivity has been reported in activity-monitoring studies
\cite{pires_improving_2020}.

A different line of work aims to learn features that remain useful under missing or noisy observations. MaskCAE \cite{cheng_maskcae_2024} learns sensor representations from masked inputs with a convolutional autoencoder. Liu \textit{et al.} \cite{liu_disentangling_2024} improve robustness to noisy signals by separating multi-scale frequency components, while Hossain \textit{et al.} \cite{hossain_method_2020} study activity recognition under missing-data conditions. These methods reduce the dependence on explicit reconstruction. However, their performance during deployment depends heavily on whether the pre-training corruptions match realistic failure topologies and whether the encoder explicitly models observation validity. MCSTN addresses these two issues by generating missingness according to the sensor hierarchy and by providing the resulting availability mask directly to the encoder. This allows the model to distinguish different failure structures even when they produce a similar overall missing rate.

\subsection{Consistency and Multi-View Learning under Missing Views}

Multi-view learning provides a natural framework for robust HAR by treating degraded sensor signals as partial views of an underlying physical activity. Consistency regularization aligns representations across differently perturbed inputs, commonly implemented through contrastive objectives or knowledge distillation \cite{xiao_federated_2025, li_feature-level_2024}. In multimodal applications with missing modalities, cross-modal translation methods synthesize missing views before feature fusion \cite{liu_modality_2024}, which reintroduces generative uncertainty. Sensor foundation models similarly employ masked or contrastive pre-training on multi-view streams \cite{haresamudram_masked_2020, cheng_maskcae_2024}.

MCSTN couples structured view generation with an asymmetric consistency target.
Student views contain point, temporal-block, axis, modality, or node failures,
with independently sampled missingness and noise. An EMA teacher encodes the
reference observation of the same window, and the student
representations are aligned with this common reference. The teacher receives
no synthetic missingness or noise, so its reference contains the information
available before student-view degradation.


\section{Methodology}
\label{sec:method}

\subsection{Overview}
\label{subsec:overview}

Let
$\mathbf{X}=[\mathbf{x}_{1},\ldots,\mathbf{x}_{L}]^{\top}
\in\mathbb{R}^{L\times C}$
denote a segmented sensor window with $L$ time steps and $C$ channels, and let
$y\in\{1,\ldots,N_{\mathrm{cls}}\}$ be its activity label.
The channels are grouped into $N$ body-worn sensor nodes
$\{\mathcal{C}_{n}\}_{n=1}^{N}$ according to the sensing topology.
Each channel belongs to one measurement axis, sensing modality, and sensor node.

Sensor measurements may be incomplete or noisy.
Missing observations can occur as isolated samples, contiguous temporal gaps,
or persistent axis-, modality-, and node-level failures.
We represent an imperfect view using an availability mask
$\mathbf{M}\in\{0,1\}^{L\times C}$, where $M_{tc}=1$ indicates that channel
$c$ is available at time $t$.
A severity variable $s\in[0,1]$ controls the noise level during training.
The mask is provided to the recognition model, whereas $s$ is used only by
the imperfection generator.

MCSTN does not reconstruct missing measurements. It learns activity representations
that remain stable across complete and imperfect observations.

\begin{figure}[!t]
\centering
\includegraphics[width=\linewidth]{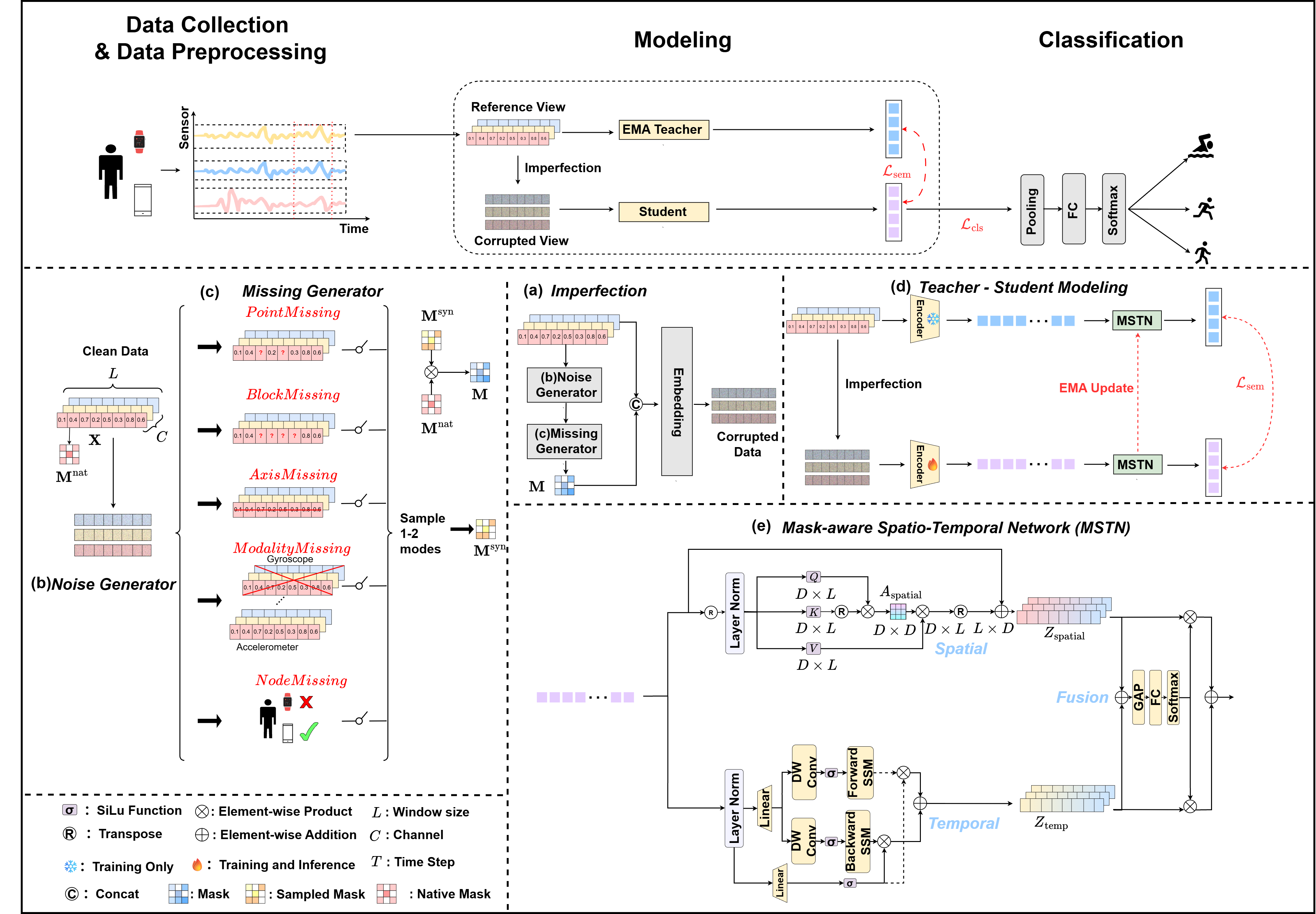}
\caption{Overview of MCSTN. The upper panel illustrates the complete pipeline
from multi-sensor data preprocessing to activity classification. 
(a) Structured imperfect views are constructed by combining noise injection
with sampled missingness. 
(b) The noise generator perturbs the observed signals according to the
corruption severity. 
(c) The missingness generator simulates point-, block-, axis-, modality-, and
node-level sensor failures while preserving native missingness. 
(d) During training, an EMA teacher processes the reference view and guides
the student representations of corrupted views through semantic consistency.
(e) MCSTN encodes each view with parallel mask-aware spatial and
bidirectional temporal branches, followed by adaptive feature fusion.
The teacher and imperfection generator are removed at inference.}
\label{fig:overview}
\end{figure}

Fig.~\ref{fig:overview} shows the three components. The structured imperfection
generator (Section~\ref{subsec:structured_imperfection}) produces training views
with different missing patterns and noise levels. The mask-aware representation
(Section~\ref{subsec:mask_representation}) encodes the availability mask together
with the measurements, and the parallel encoder
(Section~\ref{subsec:parallel_encoder}) uses it in attention and temporal
aggregation. Complete-to-imperfect learning (Sections~\ref{subsec:teacher_student}
and~\ref{subsec:semantic_consistency}) transfers semantics from the reference view
to the imperfect student views. Only the student network is kept at inference
(Section~\ref{subsec:training}).

\subsection{Structured Imperfection Modeling}
\label{subsec:structured_imperfection}

Sensor failures occur at different structural levels.
We therefore generate missing patterns according to the physical sensor
organization rather than using only entry-wise random masking.
Noise is modeled separately from missingness.

Each channel is first standardized using statistics estimated from the
training split.
Noise is then added to the standardized signal, followed by the availability
mask.
This keeps unavailable entries at zero and defines the noise level on
standardized measurements.

\subsubsection{Structured Missing Patterns}
\label{subsubsec:patterns}

We consider five failure modes,
\[
\mathcal{Q}
=
\{
q_{\mathrm{pt}},
q_{\mathrm{blk}},
q_{\mathrm{ax}},
q_{\mathrm{mod}},
q_{\mathrm{nd}}
\},
\]
corresponding to point missingness, temporal-block missingness, axis failure,
modality dropout, and node dropout.
Each mode has its own corruption level, as summarized in
Table~\ref{tab:missing_modes}.

\begin{table}[t]
\centering
\caption{Failure modes and their corruption levels.
$N_{\mathrm{mod}}$ denotes the number of (node, modality) groups.}
\label{tab:missing_modes}
\small
\begin{tabular}{@{}llll@{}}
\toprule
Mode & Corruption level & Range & Failure type \\
\midrule
$q_{\mathrm{pt}}$
& $\rho_{\mathrm{pt}}$
& $[0,\rho^{\max}_{\mathrm{pt}}]$
& isolated samples \\

$q_{\mathrm{blk}}$
& $(\rho_{\mathrm{blk}},\mathcal{B})$
& $[0,\rho^{\max}_{\mathrm{blk}}]
\times\{\text{ch.},\text{mod.},\text{node}\}$
& temporal interruption \\

$q_{\mathrm{ax}}$
& $m_{\mathrm{ax}}$
& $\{1,\ldots,C-1\}$
& axis failure \\

$q_{\mathrm{mod}}$
& $m_{\mathrm{mod}}$
& $\{1,\ldots,N_{\mathrm{mod}}-1\}$
& modality dropout \\

$q_{\mathrm{nd}}$
& $m_{\mathrm{nd}}$
& $\{1,\ldots,N-1\}$
& node dropout \\
\bottomrule
\end{tabular}
\end{table}

For each imperfect view,
$K_{\mathrm{mode}}\in\{1,2\}$ distinct modes are sampled without replacement.
For each sampled mode $q_k$, a corruption level $\rho_k$ is drawn from its
valid range and used to generate a binary availability mask,

\[
\mathbf{M}^{(k)}
=
\mathcal{G}_{q_k}(\rho_k).
\]

The sampled masks are combined by element-wise multiplication:

\begin{equation}
\mathbf{M}^{\mathrm{syn}}
=
\mathbf{M}^{(1)}
\odot
\cdots
\odot
\mathbf{M}^{(K_{\mathrm{mode}})}.
\label{eq:synthetic_mask}
\end{equation}

Mode-specific corruption levels are used because different failure types with
the same number of missing entries can remove different sensor information.

Missing values already present in the recorded data are retained through the
native mask $\mathbf{M}^{\mathrm{nat}}$.
The final availability mask is

\begin{equation}
\mathbf{M}
=
\mathbf{M}^{\mathrm{nat}}
\odot
\mathbf{M}^{\mathrm{syn}}.
\label{eq:effective_mask}
\end{equation}

Synthetic corruption cannot convert a natively missing value into an
observed one. When no native missing values are present,
$\mathbf{M}^{\mathrm{nat}}=\mathbf{1}$.

\subsubsection{Noise on Observed Measurements}
\label{subsubsec:noise}

Missingness determines which measurements are available, while noise changes
the quality of the available measurements.
Given an availability mask $\mathbf{M}$, the corrupted input is

\begin{equation}
\widetilde{\mathbf{X}}
=
\mathbf{M}
\odot
\left(
\mathbf{X}
+
\sigma(s)\boldsymbol{\epsilon}
\right),
\qquad
\boldsymbol{\epsilon}
\sim
\mathcal{N}(\mathbf{0},\mathbf{I}),
\label{eq:corruption}
\end{equation}

where the noise amplitude $\sigma(s)$ follows an SNR-based schedule. The severity $s$ is used only to generate training views and is never
provided to MCSTN.

\subsection{Mask-Aware Sensor Representation}
\label{subsec:mask_representation}

After standardization, zero is a valid sensor value.
Using zero alone to represent a missing entry makes missingness
ambiguous. MCSTN therefore encodes the availability mask together with the
sensor measurements.

For node $n$ at time step $t$, we construct

\begin{equation}
\mathbf{v}_{t,n}
=
\left[
\widetilde{\mathbf{X}}^{t,\mathcal{C}_{n}};
\mathbf{M}^{t,\mathcal{C}_{n}}
\right]
\in
\mathbb{R}^{2|\mathcal{C}_{n}|}.
\label{eq:node_input}
\end{equation}

The node representation is

\begin{equation}
\mathbf{h}^{(0)}_{t,n}
=
\mathbf{W}^{(n)}_{e}\mathbf{v}_{t,n}
+
\mathbf{b}^{(n)}_{e}
+
\mathbf{e}^{\mathrm{node}}_{n}
+
\mathbf{e}^{\mathrm{pos}}_{t}
\in
\mathbb{R}^{d},
\label{eq:node_embedding}
\end{equation}

where $\mathbf{W}^{(n)}_{e}$ and $\mathbf{b}^{(n)}_{e}$ are node-specific
projection parameters,
$\mathbf{e}^{\mathrm{node}}_{n}$ is a learnable node embedding, and
$\mathbf{e}^{\mathrm{pos}}_{t}$ is a temporal positional embedding. The node features are concatenated as

\begin{equation}
\mathbf{h}^{(0)}_{t}
=
\left[
\mathbf{h}^{(0)}_{t,1};
\ldots;
\mathbf{h}^{(0)}_{t,N}
\right]
\in
\mathbb{R}^{D},
\qquad
D=Nd.
\label{eq:embedding}
\end{equation}

Gathering all time steps gives
$\mathbf{H}^{(0)}\in\mathbb{R}^{L\times D}$.
Node-specific projections support different channel configurations across
sensor nodes, while the node embeddings retain sensor-location information.

We also compute the proportion of observed channels at each time step,

\begin{equation}
a_t
=
\frac{1}{C}
\sum_{c=1}^{C}
M_{tc},
\qquad
\mathbf{a}
=
[a_1,\ldots,a_L]^{\top}.
\label{eq:availability}
\end{equation}

The availability vector $\mathbf{a}$ is used in the transposed-attention
branch and the final temporal aggregation.

\subsection{Parallel Spatio-Temporal Encoder}
\label{subsec:parallel_encoder}

Activity recognition depends on both temporal dynamics and interactions
between sensor features. MCSTN stacks $N_b$ encoder blocks, each containing a temporal
state-space branch and a transposed-attention branch.
For block $b$,

\begin{equation}
\mathbf{H}^{(b+1)}
=
\mathrm{Block}_{b}
\left(
\mathbf{H}^{(b)},
\mathbf{a}
\right),
\qquad
b=0,\ldots,N_b-1,
\label{eq:encoder_block}
\end{equation}
and the final sequence representation is
$\mathbf{Z}=\mathbf{H}^{(N_b)}$.
The two branches use separate normalization and projection parameters.

\subsubsection{Bi-Directional Selective State-Space Branch}
\label{subsubsec:bimamba}

The temporal branch adopts a Mamba-style selective state-space model (SSM)
to capture long-range temporal dependencies with linear sequence complexity
\cite{gu_mamba_2023}.
A continuous SSM is written as
\begin{equation}
\frac{\mathrm{d}\mathbf{h}(t)}{\mathrm{d}t}
=
\mathbf{A}\mathbf{h}(t)
+
\mathbf{B}\mathbf{x}(t),
\qquad
\mathbf{y}(t)
=
\mathbf{C}\mathbf{h}(t),
\label{eq:ssm_continuous}
\end{equation}
where $\mathbf{h}(t)$ is the latent state and
$\mathbf{x}(t)$ and $\mathbf{y}(t)$ denote the input and output.
Mamba makes the discretized SSM input-dependent by conditioning its
step size and input/output projections on the current features.
We use the standard selective SSM formulation and omit its recurrence here
for brevity.

Given the embedded feature sequence $\mathbf{X}_{\mathrm{emb}}$, we apply
independent forward and backward selective scans:
\begin{equation}
\mathbf{Z}_{\mathrm{ssm}}
=
\operatorname{SSM}_{\mathrm{fwd}}
\left(
\mathbf{X}_{\mathrm{emb}}
\right)
+
\operatorname{flip}
\left[
\operatorname{SSM}_{\mathrm{bwd}}
\left(
\operatorname{flip}
\left(
\mathbf{X}_{\mathrm{emb}}
\right)
\right)
\right].
\label{eq:bidirectional_ssm}
\end{equation}

An input-dependent gate modulates the bi-directional representation:
\begin{equation}
\mathbf{G}
=
\operatorname{SiLU}
\left(
\mathbf{X}_{\mathrm{emb}}
\mathbf{W}_{g}
\right),
\qquad
\mathbf{Z}_{\mathrm{temp}}
=
\left(
\mathbf{Z}_{\mathrm{ssm}}
\odot
\mathbf{G}
\right)
\mathbf{W}_{o},
\label{eq:temporal_branch}
\end{equation}
where $\mathbf{W}_{g}$ is the gating projection and
$\mathbf{W}_{o}$ projects the gated representation to the block dimension.
The two scans use independent parameters. Scanning in both directions is
possible because the whole segmented window is available at inference.

\subsubsection{Mask-Aware Transposed Attention}
\label{subsubsec:spatial_attention}

The second branch models dependencies between latent feature dimensions.
Given the normalized input
$\mathbf{H}_{\mathrm{sp}}
=\mathrm{LN}_{\mathrm{sp}}(\mathbf{H}^{(b)})$,
we first project it into query, key, and value features and split them into
$N_h$ heads. To reduce the contribution of poorly observed time steps, the query and key
features are weighted by the availability vector $\mathbf{a}$ before
normalization. For head $h$,

\begin{equation}
\mathbf{Q}_{h}
=
\mathrm{norm}
\left[
(\mathbf{H}_{\mathrm{sp}}\mathbf{W}_{Q}^{h})^{\top}
\mathrm{diag}(\sqrt{\mathbf{a}})
\right],
\qquad
\mathbf{K}_{h}
=
\mathrm{norm}
\left[
(\mathbf{H}_{\mathrm{sp}}\mathbf{W}_{K}^{h})^{\top}
\mathrm{diag}(\sqrt{\mathbf{a}})
\right],
\label{eq:mask_qk}
\end{equation}

where
$\mathbf{Q}_{h},\mathbf{K}_{h}\in\mathbb{R}^{D_h\times L}$
and $D_h=D/N_h$. The latent attention map and branch output are then computed as

\begin{equation}
\mathbf{A}_{h}
=
\mathrm{Softmax}
\left(
\gamma
\mathbf{Q}_{h}
\mathbf{K}_{h}^{\top}
\right),
\qquad
\mathbf{Z}_{\mathrm{spatial}}
=
\mathrm{Concat}_{h}
\left[
\mathbf{A}_{h}
(\mathbf{H}_{\mathrm{sp}}\mathbf{W}_{V}^{h})^{\top}
\right]^{\top}
\mathbf{W}_{O},
\label{eq:transposed_attention}
\end{equation}

where $\gamma$ is a learnable scaling factor.
Unlike temporal self-attention,
$\mathbf{A}_{h}\in\mathbb{R}^{D_h\times D_h}$
models dependencies between latent feature dimensions rather than time steps.

\subsubsection{Adaptive Feature Fusion}
\label{subsubsec:adaptive_fusion}

The temporal and transposed-attention branches provide complementary
representations. We use input-dependent weights to combine their outputs.
Given $\mathbf{Z}_{\mathrm{temp}}$ and $\mathbf{Z}_{\mathrm{spatial}}$,
the fusion weights are computed from their global context:

\begin{equation}
[\alpha,\beta]
=
\operatorname{Softmax}
\left(
\mathbf{W}_{f}
\operatorname{GAP}
\left(
\mathbf{Z}_{\mathrm{temp}}
+
\mathbf{Z}_{\mathrm{spatial}}
\right)
\right),
\label{eq:fusion_weights}
\end{equation}

where $\operatorname{GAP}(\cdot)$ denotes global average pooling and
$\alpha$ and $\beta$ are the weights of the temporal and spatial branches,
respectively. The fused representation is

\begin{equation}
\mathbf{Z}_{\mathrm{fuse}}
=
\alpha\mathbf{Z}_{\mathrm{temp}}
+
\beta\mathbf{Z}_{\mathrm{spatial}}.
\label{eq:adaptive_fusion}
\end{equation}

\subsection{Complete-to-Imperfect Multi-View Learning}
\label{subsec:teacher_student}

Different imperfect views of the same activity may retain different sensor
information. Aligning two degraded views can therefore encourage agreement
between representations that are both incomplete. We instead use the most
complete available observation as a common reference.

The reference view is defined as
$\mathbf{X}^{\mathrm{ref}}
=\mathbf{M}^{\mathrm{nat}}\odot\mathbf{X}$.
An EMA teacher with parameters $\bar{\theta}$ processes this reference,
while the student with parameters $\theta$ receives $K_v$ independently
corrupted views generated using Eq.~\eqref{eq:corruption}.

Their representations are

\begin{equation}
\mathbf{z}^{T}
=
\mathrm{Pool}\,
f_{\bar{\theta}}
\left(
\mathbf{X}^{\mathrm{ref}},
\mathbf{M}^{\mathrm{nat}}
\right),
\qquad
\mathbf{z}^{S}_{j}
=
\mathrm{Pool}\,
f_{\theta}
\left(
\widetilde{\mathbf{X}}_{j},
\mathbf{M}_{j}
\right),
\quad
j=1,\ldots,K_v .
\label{eq:teacher_student}
\end{equation}

The teacher receives no synthetic missingness or noise, whereas the student views are generated independently with different missing patterns and noise levels. The teacher is initialized from the student and updated by EMA,
$\bar{\theta}\leftarrow
\mu\bar{\theta}+(1-\mu)\theta$.
With probability $p_{\mathrm{cl}}$, identity corruption is used for a student view to retain supervision on complete inputs.

\subsection{Semantic Consistency Learning}
\label{subsec:semantic_consistency}

Each student view is supervised by the activity label, and
$\mathcal{L}_{\mathrm{cls}}$ denotes the average cross-entropy loss over the $K_v$ views.

To keep the semantic information of the reference view in the representations of imperfect views, we align each pooled student representation with the EMA teacher representation:

\begin{equation}
\mathcal{L}_{\mathrm{sem}}
=
\frac{1}{K_v}
\sum_{j=1}^{K_v}
\left[
1-
\operatorname{cos}
\left(
\mathbf{z}^{S}_{j},
\mathrm{sg}\!\left(\mathbf{z}^{T}\right)
\right)
\right],
\label{eq:semantic_loss}
\end{equation}

where $\operatorname{cos}(\cdot,\cdot)$ denotes cosine similarity and $\mathrm{sg}(\cdot)$ denotes stop-gradient. The loss pulls the representation of every imperfect view toward the same reference representation.

The final objective is

\begin{equation}
\mathcal{L}
=
\mathcal{L}_{\mathrm{cls}}
+
\lambda_{\mathrm{sem}}(r)
\mathcal{L}_{\mathrm{sem}} .
\label{eq:total_loss}
\end{equation}

The consistency weight is linearly increased from zero to
$\lambda_{\mathrm{sem}}^{\max}$ during the first $\kappa$ fraction of training and remains fixed afterwards.

\subsection{Training and Inference}
\label{subsec:training}

During training, the EMA teacher processes one reference view and the student
processes $K_v$ independently corrupted views for each sample. The student is optimized using Eq.~\eqref{eq:total_loss}, followed by the EMA teacher update.

At inference, the imperfection generator, EMA teacher,
multi-view construction, and consistency loss are removed.
The student receives the observed sensor window and its availability mask and
performs a single forward pass for classification.
No noise-level estimate, signal reconstruction, teacher inference, or
additional view generation is required.

\section{Experiments and Results}
\label{sec:experiments}

\subsection{Datasets and Evaluation Splits}

We use three wearable HAR datasets with different sensor configurations:
PAMAP2~\cite{reiss_introducing_2012}, WISDM~\cite{kwapisz_activity_2011}, and
Opportunity~\cite{roggen_collecting_2010}. PAMAP2 records accelerometer, gyroscope, and magnetometer signals at three body locations. Opportunity has a denser body-worn network and contains substantial missing measurements in the original recordings; in the main experiments these native gaps are linearly interpolated before synthetic corruption, so $\mathbf{M}^{\mathrm{nat}}=\mathbf{1}$, and Section~\ref{sec:native} retains them. WISDM is recorded with a single pocket-worn device. Table~\ref{tab:dataset_summary} summarises the datasets and evaluation protocols.

\begin{table}[!ht]
\centering
\caption{Dataset characteristics and evaluation protocols.}
\label{tab:dataset_summary}
\resizebox{\textwidth}{!}{%
\begin{tabular}{l|ccc}
\toprule
\textbf{Parameter}
& \textbf{PAMAP2}
& \textbf{WISDM}
& \textbf{Opportunity} \\
\midrule
Positions
& Wrist, Chest, Ankle
& Pocket
& Full Body \\

Sensor Type
& A, G, M
& A
& A, G, M \\

Sampling Rate
& 33.3 Hz
& 20 Hz
& 30 Hz \\

Subject Count
& 9
& 29
& 4 \\

Activity Classes
& 12
& 6
& 17 \\

Window Size (samples)
& 171
& 200
& 30 \\

Overlap
& 50\%
& 50\%
& 50\% \\

Partition Protocol
& Subject-wise hold-out
& Subject-wise split
& Subject-and-run split \\

Training
& S1--4, S7--8
& 20 subjects
& S1/S4 R1--5; S2/S3 R1--2 \\

Validation
& S5
& 3 subjects
& S2/S3 R3 \\

Test
& S6
& 6 subjects
& S2/S3 R4--5 \\

\bottomrule
\end{tabular}%
}
\end{table}

The sensor configuration determines which failure modes can be applied to each dataset. Opportunity contains $C=113$ channels from $N=19$ sensor nodes and $39$ (node, modality) groups. PAMAP2 contains $C=36$ channels from $N=3$ nodes and $12$ groups. WISDM contains only $C=3$ channels from a single sensor node and forms one (node, modality) group. Therefore, node and modality dropout are
not considered for WISDM. Its imperfection grid includes only point, temporal-block, and axis failures, while all five failure modes are evaluated on Opportunity and PAMAP2.

The splits follow Table~\ref{tab:dataset_summary}: subject-wise hold-out for PAMAP2 and WISDM, and the run-based split of Opportunity, in which runs R4--5 of subjects S2 and S3 form the test set.

For all three datasets, the data split is performed before sliding-window
segmentation. This prevents overlapping or temporally adjacent windows from the
validation or test partitions from appearing in the training set. Channel-wise
standardisation statistics are computed from the training set only and then
applied unchanged to the validation and test sets.

\subsection{Implementation Details}

All experiments are implemented in PyTorch and run on a single NVIDIA A100
80\,GB GPU. MCSTN uses two parallel spatio-temporal blocks with a total hidden
width of approximately 192, an SSM state size of 16, four attention heads, and
a dropout rate of 0.1. Models are trained for 60 epochs with AdamW, a batch
size of 64, and an initial learning rate of $2\times10^{-3}$. The multi-view objective uses $K_v=2$ student views,
an EMA momentum of $\mu=0.996$, and
$\lambda_{\mathrm{sem}}^{\max}=1.0$. 

\subsection{Evaluation Protocol}

Macro-F1 is used as the primary metric because the three HAR datasets are
class-imbalanced. Accuracy is additionally reported in
Table~\ref{tab:sota} for comparison with prior work. With $TP_k$, $FP_k$, and
$FN_k$ denoting the per-class counts over $N_{\mathrm{cls}}$ classes,

\begin{equation}
\text{Accuracy} =
\frac{\sum_k TP_k}{\sum_k (TP_k + FN_k)},
\qquad
F1_k =
\frac{2TP_k}{2TP_k + FP_k + FN_k},
\end{equation}

\begin{equation}
\text{Macro-F1} =
\frac{1}{N_{\mathrm{cls}}}
\sum_{k=1}^{N_{\mathrm{cls}}} F1_k .
\end{equation}

Robustness is evaluated using three fixed severity levels for each applicable
failure mode. Clean, individual-failure, and mixed conditions are generated
with fixed random seeds so that all methods are evaluated on the same
corrupted observations. We report four robustness summaries. \emph{Avg.\ Imp.} is the mean macro-F1 over the applicable single-mode conditions and the mixed condition, this  are reported at the intermediate 30\% missing ratio.
\emph{Worst} is the lowest macro-F1 among all single-mode severity levels,
\emph{Mixed} is the score under combined failures, and \emph{Drop} is the
difference between the clean score and Avg.\ Imp. 

We compare MCSTN with LSTM~\cite{zeng_understanding_2018},
DeepConvLSTM~\cite{ordonez_deep_2016},
Transformer~\cite{haresamudram_masked_2020},
GRU-D~\cite{che_recurrent_2018},
BRITS~\cite{cao_brits_2018},
SAITS~\cite{du_saits_2023}, and
HARMamba~\cite{li_harmamba_2024}. For BRITS and SAITS, the imputed sequences
are passed to the same DeepConvLSTM-style classifier. All methods use the same
data splits, preprocessing, optimisation settings, and training budget. All baselines are trained with the same structured imperfection generator as MCSTN, so that differences between methods do not come from exposure to different corruption patterns during training.

\subsection{Recognition under Structured Sensor Failures}
\label{subsec:performance}

Table~\ref{tab:compare} reports the per-mode results on Opportunity, which has the largest number of channels and native missingness among the three datasets.

\begin{table}[t]
\centering
\caption{Macro-F1 (\%) under clean and imperfect sensing conditions on the Opportunity dataset. Results are reported as mean$\pm$standard deviation over
repeated runs. \textbf{Bold} denotes the best result.}
\label{tab:compare}
\scriptsize
\setlength{\tabcolsep}{2.4pt}
\renewcommand{\arraystretch}{1.08}

\begin{adjustbox}{max width=\textwidth}
\begin{tabular}{lcccccccc}
\toprule
Method
& Clean
& Point
& Block
& Axis
& Mod.
& Node
& Mixed
& Avg. Imp. \\
\midrule

LSTM
& $80.71_{\pm3.40}$
& $79.28_{\pm2.67}$
& $79.54_{\pm3.12}$
& $73.89_{\pm2.74}$
& $74.17_{\pm3.31}$
& $75.21_{\pm2.52}$
& $75.71_{\pm2.47}$
& 76.30 \\

DeepConvLSTM
& $83.64_{\pm2.40}$
& $82.14_{\pm2.89}$
& $81.66_{\pm2.73}$
& $76.55_{\pm2.24}$
& $77.23_{\pm2.43}$
& $77.79_{\pm2.41}$
& $78.72_{\pm2.68}$
& 79.02 \\

Transformer
& $85.52_{\pm1.04}$
& $83.96_{\pm2.03}$
& $83.36_{\pm1.89}$
& $77.18_{\pm2.11}$
& $78.10_{\pm2.39}$
& $79.94_{\pm1.48}$
& $81.57_{\pm1.21}$
& 80.69 \\

GRU-D
& $90.02_{\pm0.19}$
& $89.16_{\pm0.22}$
& $88.41_{\pm0.15}$
& $87.70_{\pm0.20}$
& $86.41_{\pm0.60}$
& $86.18_{\pm0.23}$
& $86.79_{\pm0.36}$
& 87.44 \\

BRITS
& $90.21_{\pm0.25}$
& $89.29_{\pm0.31}$
& $88.66_{\pm0.27}$
& $87.84_{\pm0.27}$
& $88.68_{\pm0.17}$
& $86.04_{\pm0.16}$
& $86.97_{\pm0.38}$
& 87.91 \\

HARMamba
& $90.97_{\pm0.16}$
& $89.55_{\pm0.23}$
& $89.45_{\pm0.14}$
& $87.24_{\pm0.35}$
& $86.55_{\pm0.36}$
& $86.25_{\pm0.23}$
& $87.85_{\pm0.08}$
& 87.82 \\

\midrule
\textsc{MCSTN} (Ours)
& $\mathbf{93.70_{\pm0.19}}$
& $\mathbf{93.43_{\pm0.18}}$
& $\mathbf{92.71_{\pm0.21}}$
& $\mathbf{90.32_{\pm0.17}}$
& $\mathbf{91.57_{\pm0.17}}$
& $\mathbf{90.32_{\pm0.21}}$
& $\mathbf{91.69_{\pm0.27}}$
& \textbf{91.67} \\

\bottomrule
\end{tabular}
\end{adjustbox}
\end{table}

Across the baselines, point and temporal-block missingness cause small drops, whereas axis, modality, and node failures are more damaging. For example, the Transformer drops by $1.56$ and $2.16$ macro-F1 points under point and block missingness, respectively, but by $8.34$, $7.42$, and $5.58$ points under axis, modality, and node failures. This suggests that structured group failures remove more useful sensor information than sparse or local temporal missingness. Methods designed for incomplete time series improve robustness but remain sensitive to structured failures. GRU-D and BRITS achieve imperfect averages of $87.44$ and $87.91$, compared with $80.69$ for the Transformer. However, their performance still decreases from point missingness to node failure. These methods model missingness at the variable and time-step level but do not explicitly use the sensor topology to distinguish axis-, modality-, and node-level failures. HARMamba follows the same trend: it achieves the strongest clean result among the baselines ($90.97$), but falls to $86.25$ under node failure.

MCSTN achieves the best result in every condition. It reaches $93.70$ on clean input and an imperfect average of $91.67$, exceeding the strongest baseline by $2.73$ and $3.76$ points, respectively. Its performance is also more stable across failure modes. The gap between its strongest and weakest failure modes is $3.11$ points, compared with $6.78$ the Transformer and $3.30$ HARMamba, and its clean-to-imperfect drop is $2.03$ points. Because all baselines are trained with the same imperfection generator, this gap does not come from the training corruption itself; it reflects the mask-aware encoder and complete-to-imperfect learning.

\subsection{Cross-Dataset Comparison and Computational Cost}

Table~\ref{tab:main_comparison} compares the methods across all three datasets under clean and imperfect observations.

\begin{table}[t]
\centering
\caption{Cross-dataset comparison under clean and imperfect observations.
All performance values are macro-F1 (\%).}
\label{tab:main_comparison}

\scriptsize
\setlength{\tabcolsep}{2.0pt}
\renewcommand{\arraystretch}{1.05}

\resizebox{\linewidth}{!}{%
\begin{tabular}{llccccccc}
\toprule
Dataset & Method
& Clean
& Avg. Imp.
& Worst
& Mixed
& Drop $\downarrow$
& Params (M)
& FLOPs (G) \\
\midrule

\multirow{8}{*}{OPPORTUNITY}
& LSTM
& 80.71 & 76.30 & 70.02 & 75.71 & 4.41 & 0.75 & 0.02 \\

& DeepConvLSTM
& 83.64 & 79.02 & 73.31 & 78.72 & 4.62 & 0.89 & 0.03 \\

& Transformer
& 85.52 & 80.69 & 77.12 & 81.57 & 4.83 & 0.82 & 0.04 \\

& GRU-D
& 90.02 & 87.44 & 85.95 & 86.79 & 2.58 & 0.78 & 0.03 \\

& BRITS
& 90.21 & 87.91 & 85.88 & 86.97 & 2.30 & 0.90 & 0.05 \\

& SAITS
& 87.37 & 80.89 & 74.39 & 83.00 & 6.48 & 2.21 & 0.04 \\

& HARMamba
& 90.97 & 87.82 & 86.08 & 87.85 & 3.15 & 1.12 & 0.04 \\

& \textbf{MCSTN}
& \textbf{93.70}
& \textbf{91.67}
& \textbf{90.32}
& \textbf{91.69}
& \textbf{2.03}
& 0.91 & 0.05 \\

\midrule

\multirow{8}{*}{PAMAP2}
& LSTM
& 86.58 & 85.41 & 82.27 & 85.17 & 1.17 & 0.38 & 0.13 \\

& DeepConvLSTM
& 89.28 & 88.55 & 85.97 & 87.95 & 0.73 & 0.83 & 0.14 \\

& Transformer
& 89.56 & 88.91 & 87.39 & 88.68 & 0.65 & 0.80 & 0.15 \\

& GRU-D
& 92.42 & 91.75 & 90.68 & 91.60 & 0.67 & 0.40 & 0.13 \\

& BRITS
&93.15   & 92.84 & 92.03 & 92.42 & 0.31 & 0.43 & 0.15 \\

& SAITS
& 89.80 & 89.49 & 88.11 & 88.89 & 0.31 & 2.08 & 0.16 \\

& HARMamba
& 91.30 & 91.51 & 90.43 & 91.50
& \textbf{$-0.21$}
& 0.90 & 0.09 \\

& \textbf{MCSTN}
& \textbf{94.88} & \textbf{94.24} & \textbf{92.69} & \textbf{92.71} & 0.64 & 0.79 & 0.13 \\

\midrule

\multirow{8}{*}{WISDM}
& LSTM
& 94.32 & 89.88 & 82.36 & 86.14 & 4.44 & 0.30 & 0.11 \\

& DeepConvLSTM
& 94.40 & 89.97 & 82.47 & 87.45 & 4.43 & 0.74 & 0.12 \\

& Transformer
& 95.07 & 92.96 & 87.34 & 90.61 & \textbf{2.11} & 0.65 & 0.14 \\

& GRU-D
& 96.08 & 93.23 & 89.66 & 91.57 & 2.85 & 0.32 & 0.12 \\

& BRITS
& 95.12 & 92.81 & 90.34 & 91.23 & 2.31 & 0.32 & 0.12 \\

& SAITS
& 94.69 & 90.40 & 84.94 & 87.51 & 4.29 & 2.03 & 0.15 \\

& HARMamba
& 97.83
& 95.00
& 89.75
& 93.56
& 2.83
& 0.87 & 0.10 \\

& \textbf{MCSTN}
& \textbf{98.52}
& \textbf{96.22}
& \textbf{91.89}
& \textbf{94.50}
& 2.30
& 0.69 & 0.13 \\

\bottomrule
\end{tabular}%
}
\end{table}

MCSTN achieves the highest clean and imperfect performance on all three datasets. Compared with the strongest baseline, its Avg. Imp. improves by $3.76$ points on Opportunity, $1.40$ points on PAMAP2, and $1.22$ points on
WISDM. The margin also holds in the \emph{Worst} column, where MCSTN exceeds the best baseline by $4.07$, $0.66$, and $1.55$ points on the three datasets.

The effect of missingness varies with the sensor configuration. On Opportunity, MCSTN drops by $2.03$ from clean to Avg. Imp., compared with $2.30$ for BRITS and $3.15$ for HARMamba. The drops are smaller on PAMAP2 for all methods. HARMamba even scores slightly higher under imperfect input than on clean input, although its clean score remains $3.58$ points below MCSTN. WISDM has a single sensor node and therefore no cross-node redundancy; MCSTN still achieves the highest Avg. Imp. score of $96.22$, although its clean-to-imperfect drop is similar to that of the stronger baselines. Under mixed corruption, MCSTN achieves $91.69$ on Opportunity, $92.71$ on PAMAP2, and $94.50$ on WISDM, the best result among all compared methods on each dataset, so its advantage persists when failure modes are combined.

\begin{table}[t]
    \centering
    \caption{Computational efficiency on the Opportunity dataset.}
    \label{tab:efficiency}
    \small
    \setlength{\tabcolsep}{3.5pt}
    \resizebox{\columnwidth}{!}{%
    \begin{tabular}{@{}lcccc@{}}
        \toprule
        Method & Params (M) & FLOPs (G) & Memory (GiB) & Latency (ms) \\
        \midrule
        LSTM           & 0.75 & 0.02 & 0.011 & 0.69  \\
        DeepConvLSTM   & 0.89 & 0.03 & 0.013 & 0.72  \\
        Transformer    & 0.82 & 0.04 & 0.013 & 4.58  \\
        GRU-D          & 0.78 & 0.03 & 0.011 & 11.75 \\
        BRITS          & 0.90 & 0.05 & 0.011 & 29.38 \\
        HARMamba       & 1.12 & 0.04 & 0.012 & 3.62  \\
        \midrule
        \textbf{MCSTN} & 0.91 & 0.05 & 0.012 & 3.78  \\
        \bottomrule
    \end{tabular}%
    }
\end{table}

Table~\ref{tab:efficiency} shows that the robustness gains do not require a large increase in computational cost. MCSTN has $0.91$\,M parameters and an inference latency of $3.78$\,ms per sample. Its latency is close to HARMamba
($3.62$\,ms) and lower than the Transformer ($4.58$\,ms). GRU-D and BRITS are slower at inference, requiring $11.75$\,ms and $29.38$\,ms per sample, respectively.

\subsection{Robustness across Corruption Intensity}

\begin{figure}[!t]
\centering
\includegraphics[width=\linewidth]{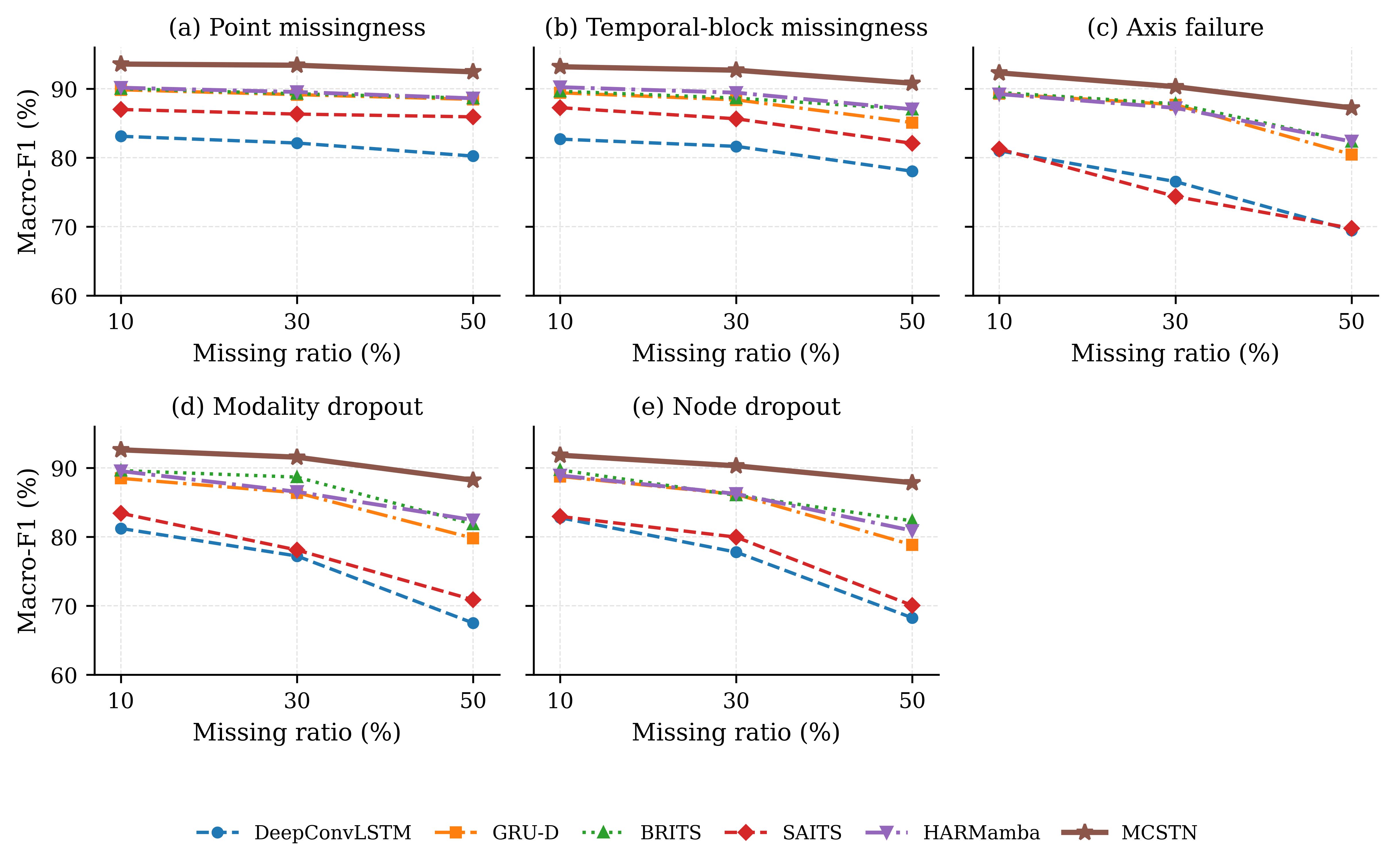}
\caption{Macro-F1 on Opportunity as a function of the missing ratio, shown separately for (a) point missingness, (b) temporal-block missingness, (c) axis failure, (d) modality dropout and (e) node dropout.}
\label{fig:severity}
\end{figure}

Fig.~\ref{fig:severity} further evaluates robustness on Opportunity by varying the missing ratio from $10\%$ to $50\%$ for each mode separately. MCSTN achieves the highest macro-F1 across all failure modes and severity levels. Point and temporal-block missingness cause relatively small performance changes for most methods, whereas axis, modality, and node failures lead to larger degradation as the missing ratio increases. From $10\%$ to $50\%$ missingness, DeepConvLSTM and SAITS lose approximately $11.5\%$--$14\%$ macro-F1 points under the structured failure modes. The stronger baselines show smaller declines, while MCSTN drops by about $4.0$ points under node failure and $5.1$ points under axis failure.

The performance gap between MCSTN and the strongest baseline also increases at higher missing ratios, particularly for modality and node failures. Axis failure remains the most challenging condition for MCSTN. Unlike node failure, which removes complete sensor nodes, axis failure removes channels across multiple nodes and therefore affects a broader part of the sensor representation.

\subsection{Comparison with State-of-the-Art Methods}

Table~\ref{tab:sota} compares MCSTN with recently published HAR methods under clean input. The results of prior methods are taken from the original papers, whose splits and window settings may differ from ours, so the table indicates competitiveness rather than a controlled comparison.

\begin{table}[!ht]
    \centering
    \caption{Clean-input accuracy and F1 score of MCSTN and published HAR methods; values of prior methods are those reported in the original papers. Best results are \textbf{boldfaced}, and the second-best results are \underline{underlined}.}
    \label{tab:sota}
    \setlength{\tabcolsep}{4pt}
    \renewcommand{\arraystretch}{1.3}
    \scriptsize
    \resizebox{\textwidth}{!}{
    \begin{tabular}{l | l | c c c c c c}
        \toprule
        \textbf{Dataset} & \multicolumn{1}{l}{} & \multicolumn{6}{c}{\textbf{SOTA Methods}} \\
        \midrule

        \multirow{3}{*}{Opportunity}
            & Methods
                & Yu \textit{et al.} \cite{yu_human_2024}
                & Zhang \textit{et al.} \cite{zhang_improved_2023}
                & Wang \textit{et al.} \cite{wang_patchhar_2024}
                & Yao \textit{et al.} \cite{yao_revisiting_2024}
                & Al-qaness \textit{et al.} \cite{al-qaness_multi-resatt_2022}
                & \textbf{MCSTN (Ours)} \\
            & Accuracy (\%)
                & 87.48
                & 90.80
                & \underline{91.11}
                & 87.96
                & 87.82
                & \textbf{94.39} \\
            & F1 Score (\%)
                & 87.00
                & 90.85
                & \underline{91.75}
                & 87.88
                & 86.37
                & \textbf{93.70} \\

        \midrule

        \multirow{3}{*}{PAMAP2}
            & Methods
                & Teng \textit{et al.} \cite{teng_innovative_2024}
                & Yu \textit{et al.} \cite{yu_human_2024}
                & Wang \textit{et al.} \cite{wang_robust_2024}
                & Huang \textit{et al.} \cite{huang_channel-equalization-har_2022}
                & Essa \textit{et al.} \cite{essa_temporal-channel_2023}
                & \textbf{MCSTN (Ours)} \\
            & Accuracy (\%)
                & \underline{94.34}
                & 89.72
                & 92.89
                & 92.14
                & 89.10
                & \textbf{94.96} \\
            & F1 Score (\%)
                & \underline{94.35}
                & 90.00
                & 93.06
                & 92.18
                & 86.42
                & \textbf{94.88} \\

        \midrule

        \multirow{3}{*}{WISDM}
            & Methods
                & Xiao \textit{et al.} \cite{xiao_federated_2025}
                & Li \textit{et al.} \cite{li_p2lhap_2024}
                & Liang \textit{et al.} \cite{liang_collaborative_2023}
                & Li \textit{et al.} \cite{li_multiresolution_2023}
                & Yi \textit{et al.} \cite{yi_human_2023}
                & \textbf{MCSTN (Ours)} \\
            & Accuracy (\%)
                & -
                & 97.51
                & \underline{97.59}
                & 97.11
                & 96.07
                & \textbf{98.60} \\
            & F1 Score (\%)
                & 89.96
                & 97.52
                & 97.52
                & \textbf{98.64}
                & 96.14
                & \underline{98.52} \\

        \bottomrule
    \end{tabular}
    }
\end{table}

On Opportunity, MCSTN achieves $94.39\%$ accuracy and $93.70\%$ macro-F1, outperforming the strongest previous result~\cite{wang_patchhar_2024} by $3.28$ and $1.95$ points, respectively. On PAMAP2, it reaches $94.96\%$
accuracy and $94.88\%$ macro-F1, improving over \cite{teng_innovative_2024} by $0.62$ and $0.53$ points. On WISDM, MCSTN obtains the highest accuracy of $98.60\%$, $1.01$ points above \cite{liang_collaborative_2023}. Its macro-F1 of $98.52\%$ is the second-best result, only $0.12$ points below \cite{li_multiresolution_2023}.

MCSTN is designed for incomplete observations and remains competitive on clean input; its main advantage appears under structured sensor failures.

\subsection{Ablation Study}
\label{subsec:ablation}

Table~\ref{tab:ablation_study} evaluates the contribution of each MCSTN component on Opportunity.

\begin{table}[t]
\centering
\caption{Ablation results on Opportunity in macro-F1 (\%).}
\label{tab:ablation_study}

\scriptsize
\setlength{\tabcolsep}{2.2pt}
\renewcommand{\arraystretch}{1.05}

\resizebox{\linewidth}{!}{%
\begin{tabular}{clcccccccc}
\toprule
ID & Configuration
& Clean
& Point
& Block
& Axis
& Mod.
& Node
& Mixed
& Avg. Imp. \\
\midrule

A0 & Full MCSTN
& 93.70 & 93.43 & 92.71 & 90.32 & 91.57 & 90.32 & 91.69 & 91.67 \\

A1 & w/o structured imperfection modeling
& 92.74 & 89.71 & 90.10 & 87.53 & 86.89 & 86.72 & 88.05 & 88.17 \\

A2 & w/o mask-aware representation
& 93.37 & 93.04 & 92.48 & 85.95 & 83.19 & 90.20 & 90.45 & 89.22 \\

A3 & w/o latent transposed-attention branch
& 92.77 & 92.02 & 91.40 & 88.54 & 88.61 & 88.70 & 89.53 & 89.80 \\

A4 & w/o bi-directional SSM branch
& 91.08 & 90.82 & 89.77 & 87.26 & 85.84 & 87.61 & 88.58 & 88.31 \\

A5 & w/o adaptive branch fusion
& 93.25 & 92.89 & 92.28 & 88.23 & 89.82 & 89.97 & 90.12 & 90.55 \\

A6 & w/o complete-to-imperfect learning
& 93.06 & 92.69 & 92.25 & 86.21 & 82.83 & 89.96 & 90.24 & 89.03 \\

\bottomrule
\end{tabular}%
}
\end{table}

The two encoder branches contribute to both clean and imperfect recognition.
Removing the bi-directional SSM branch (A4) causes the largest clean-data drop
of any component, from 93.70 to 91.08, and reduces Avg.~Imp.\ by 3.36 points. Removing the transposed-attention branch (A3)
reduces the clean score by 0.93 points and Avg.~Imp.\ by 1.87 points. Replacing
adaptive fusion with a fixed combination (A5) lowers Avg.~Imp.\ by 1.12 points.

The components designed for incomplete observations have a larger effect under
imperfect inputs than on clean data. Removing structured imperfection modelling
(A1) reduces the clean score by only 0.96 points, but lowers Avg.~Imp.\ by 3.50
points, from 91.67 to 88.17. The reduction appears in every failure mode,
ranging from 2.61 points under temporal-block missingness to 4.68 points under
modality dropout.

Removing the mask-aware representation (A2) or complete-to-imperfect learning
(A6) leaves the clean, point, block, and node scores almost unchanged (at most
0.74 points), but causes large losses under axis failure (4.37 and 4.11 points)
and modality dropout (8.38 and 8.74 points), reducing Avg.~Imp.\ by 2.45 and
2.64 points, respectively. These two failure types remove only part of the
channels within a node, so explicit availability information and a complete
reference view are most useful in this case. Node failure is less affected,
possibly because a fully missing node already produces a distinctive input
pattern through its node-specific projection and embedding.

\subsection{Structured Augmentation for Competing Architectures}

\begin{figure}[!t]
\centering
\includegraphics[width=0.62\linewidth]{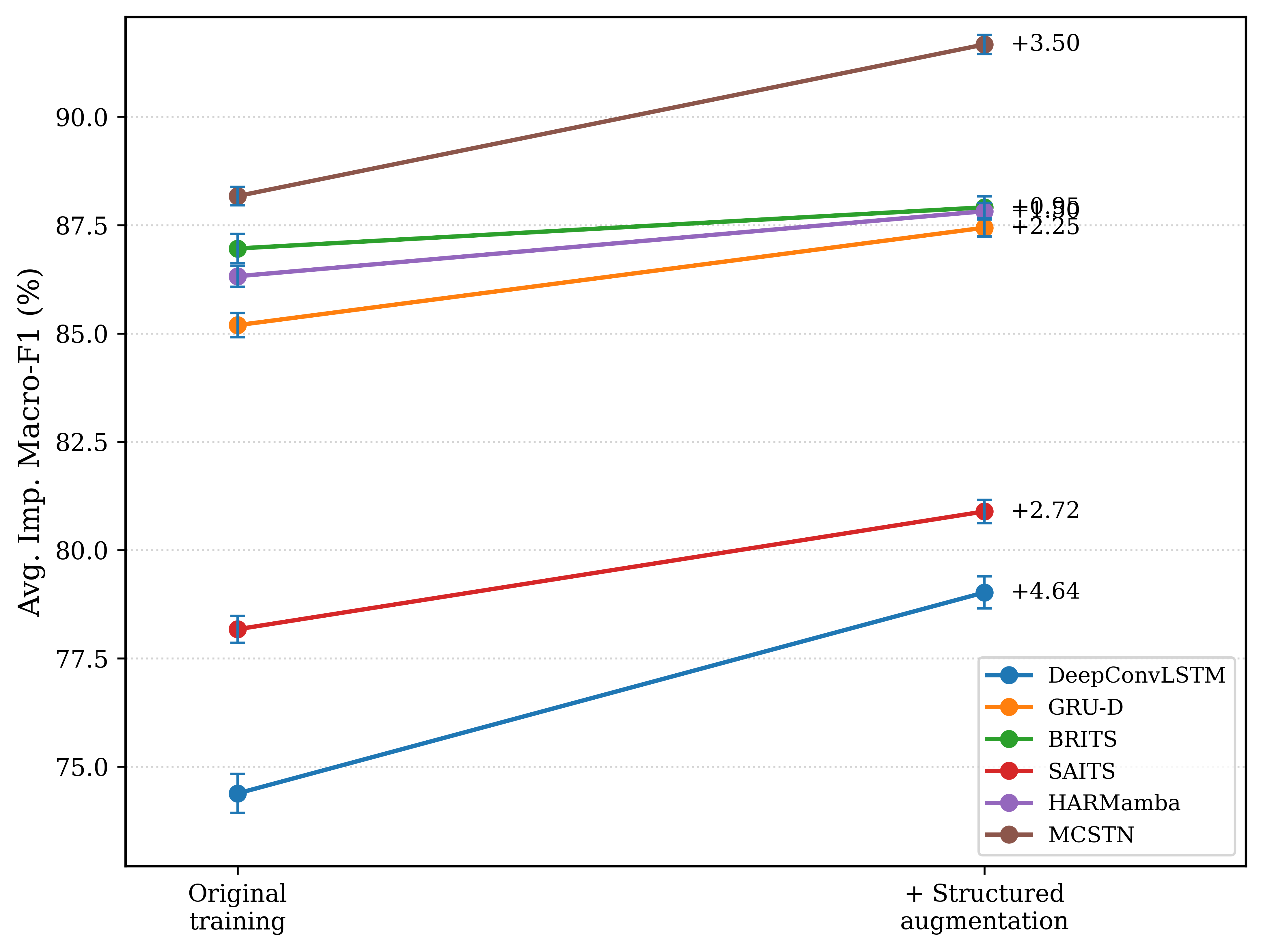}
\caption{Effect of structured imperfection augmentation on Opportunity.}
\label{fig:aug_transfer}
\end{figure}

Fig.~\ref{fig:aug_transfer} compares each method trained with and without the structured imperfection generator. The generator improves Avg.\ Imp.\ for every evaluated method, by 0.95 points for BRITS up to 4.64 points for DeepConvLSTM, so its benefit is not specific to MCSTN.

The generator alone does not close the performance gap. After augmentation, the strongest baseline (BRITS) reaches 87.91, compared with 91.67 for MCSTN, and MCSTN gains 3.50 points from the generator, more than the two strongest baselines, HARMamba (1.50) and BRITS (0.95).

\subsection{Sensitivity to the Number of Student Views}

\begin{figure}[!t]
\centering
\includegraphics[width=\linewidth]{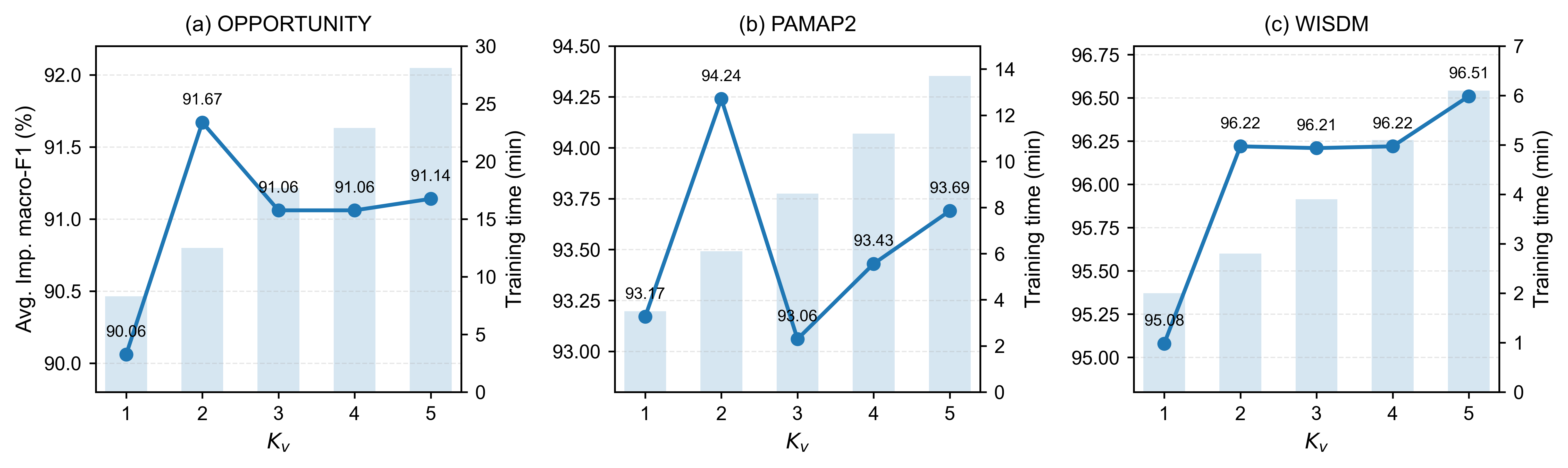}
\caption {Average imperfect macro-F1 and training time for different numbers of student views $K_v$ on Opportunity, PAMAP2, and WISDM.}
\label{fig:kv}
\end{figure}

The number of student views $K_v$ affects both training diversity and computational cost. Fig.~\ref{fig:kv} varies $K_v$ from $1$ to $5$ on all three datasets.

Increasing $K_v$ from $1$ to $2$ improves average imperfect macro-F1 by $1.61$, $1.07$, and $1.14$ points on Opportunity, PAMAP2, and WISDM, respectively. Beyond $K_v=2$, macro-F1 falls by 0.53--0.61 points on Opportunity and by 0.55--1.18 points on PAMAP2, and changes by at most 0.29 points on WISDM, while training time grows roughly linearly with $K_v$. We therefore use $K_v=2$ in the main experiments.

\subsection{Evaluation under Native Missingness}
\label{sec:native}

\begin{figure}[!t]
\centering
\includegraphics[width=\linewidth]{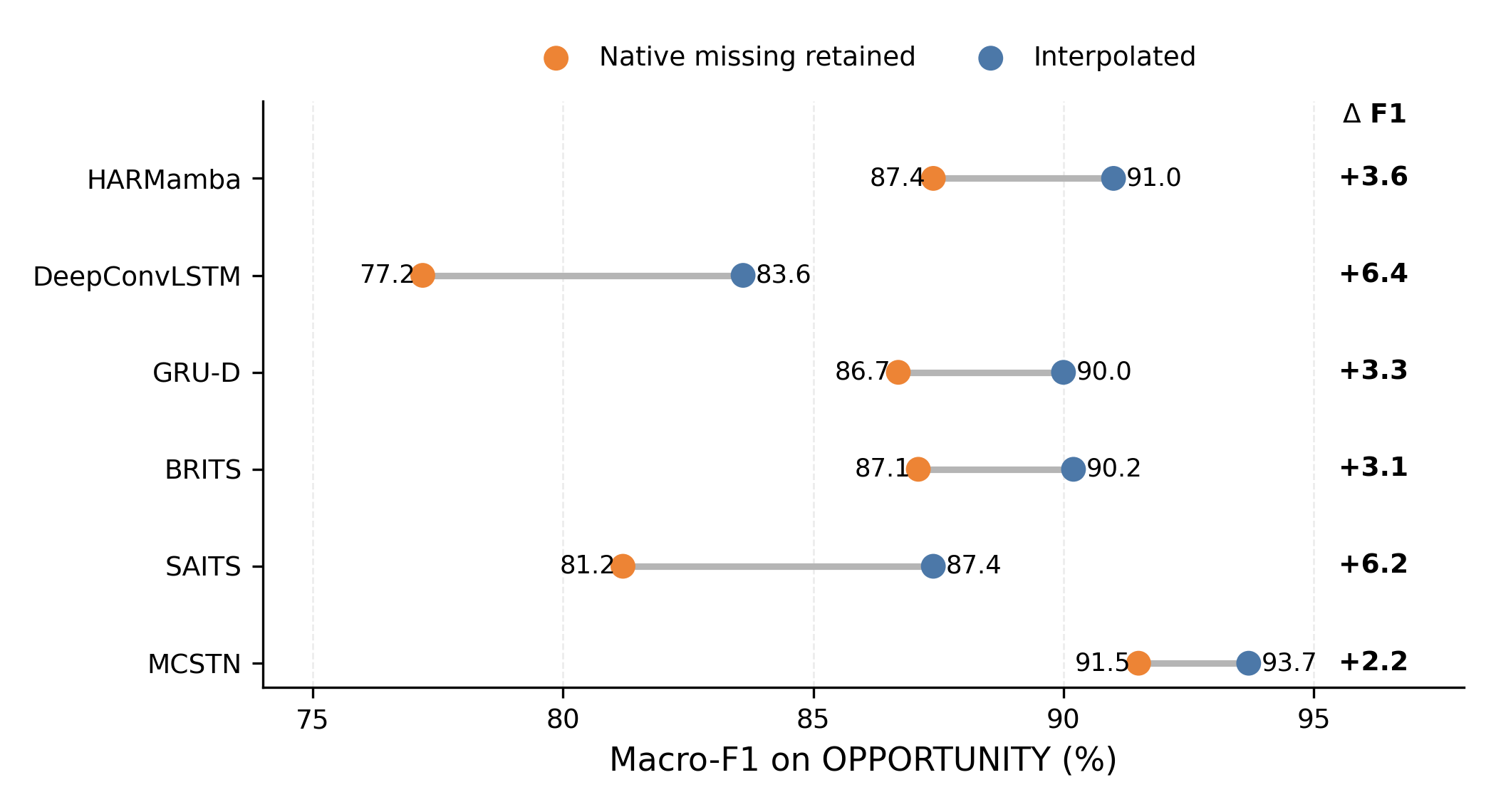}
\caption {Clean macro-F1 on Opportunity with native missing values interpolated
or retained.}
\label{fig:native}
\end{figure}

The preceding experiments add synthetic corruption to Opportunity after its
native gaps have been linearly interpolated, following common practice.
Interpolation, however, is itself a form of reconstruction and may hide the
effect of real sensor dropouts. We therefore retain the native missing values,
encode them in $\mathbf{M}^{\mathrm{nat}}$, and retrain all methods under
otherwise identical settings (Fig.~\ref{fig:native}).

Retaining native gaps degrades all methods, but to very different extents.
DeepConvLSTM and SAITS are the most affected, losing 6.4 and 6.2 points,
whereas GRU-D and BRITS, which model missingness explicitly, lose 3.3 and 3.1
points. MCSTN shows the smallest reduction (2.2 points) and reaches 91.5
macro-F1, 4.1 points above the strongest baseline (HARMamba, 87.4). This margin
is larger than under interpolation (2.7 points), indicating that the advantage
of MCSTN grows when real sensor gaps are not removed by preprocessing. These
results also show that the treatment of native missing values substantially
affects reported performance on Opportunity and should be stated explicitly in
the evaluation protocol.

\subsection{Qualitative Analysis}

\begin{figure}[!ht]
    \centering

    \begin{subfigure}[t]{0.24\linewidth}
        \centering
        \includegraphics[width=\linewidth]{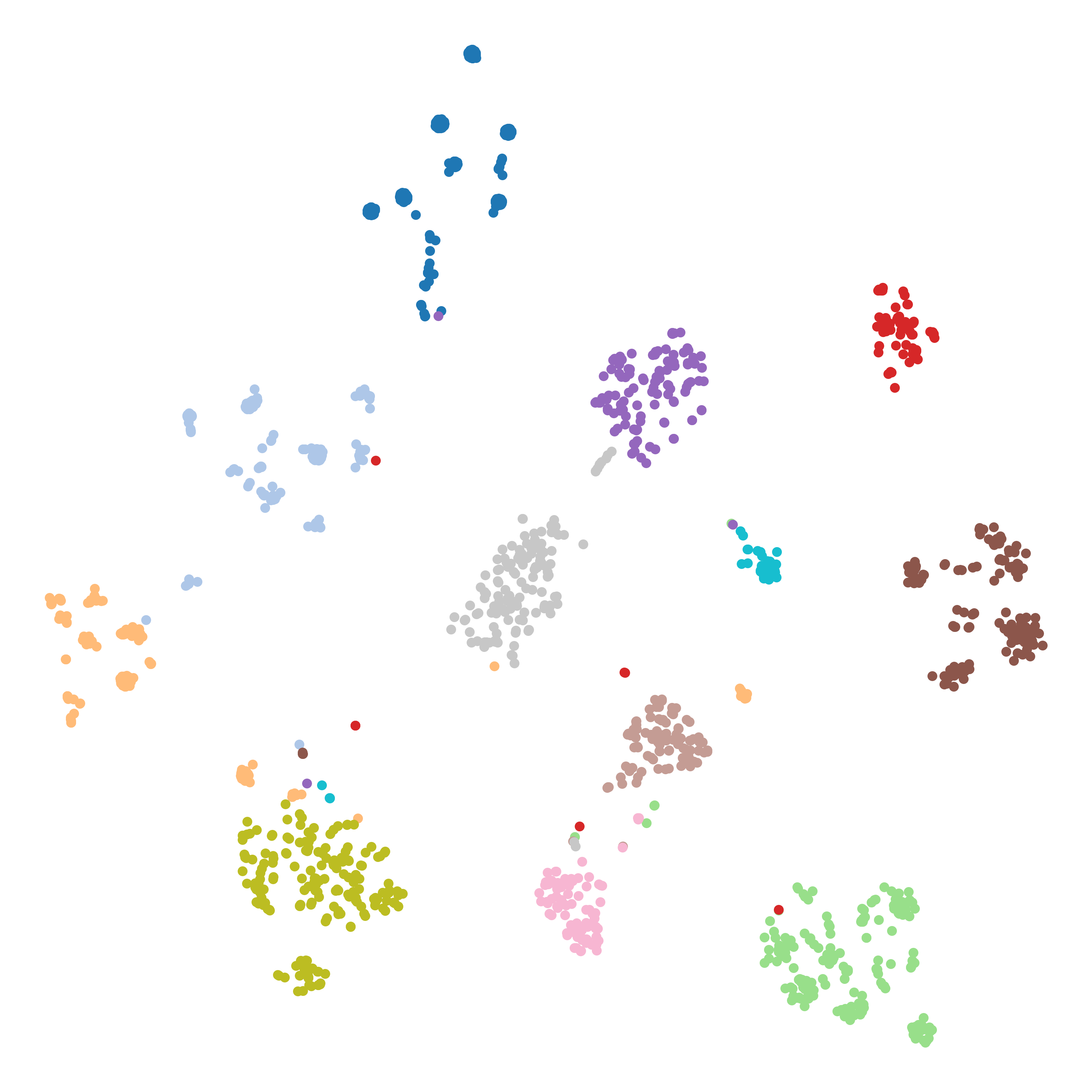}
        \caption{MCSTN -- Clean}
        \label{fig:tsne_mcstn_clean}
    \end{subfigure}\hfill
    \begin{subfigure}[t]{0.24\linewidth}
        \centering
        \includegraphics[width=\linewidth]{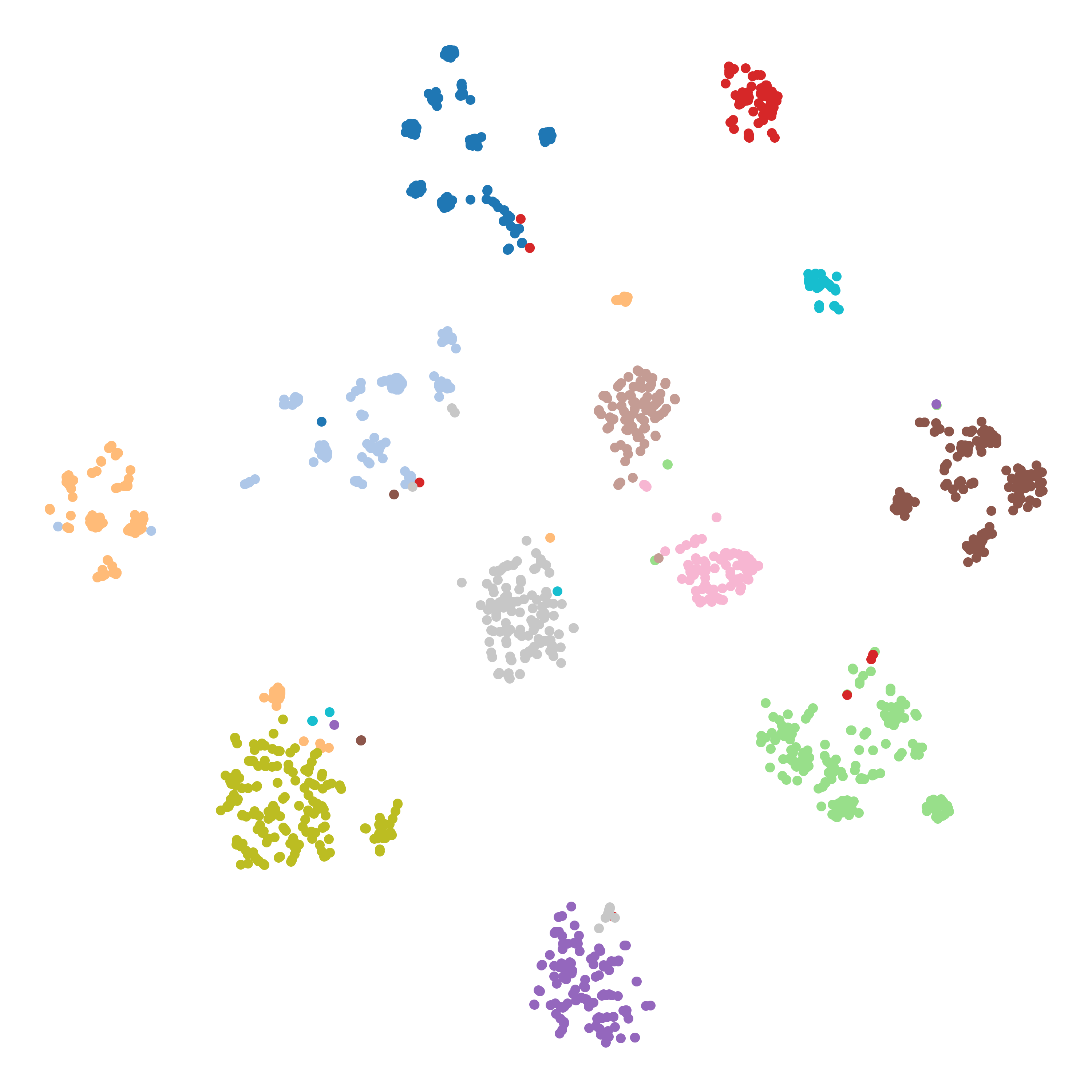}
        \caption{MCSTN -- Light}
        \label{fig:tsne_mcstn_light}
    \end{subfigure}\hfill
    \begin{subfigure}[t]{0.24\linewidth}
        \centering
        \includegraphics[width=\linewidth]{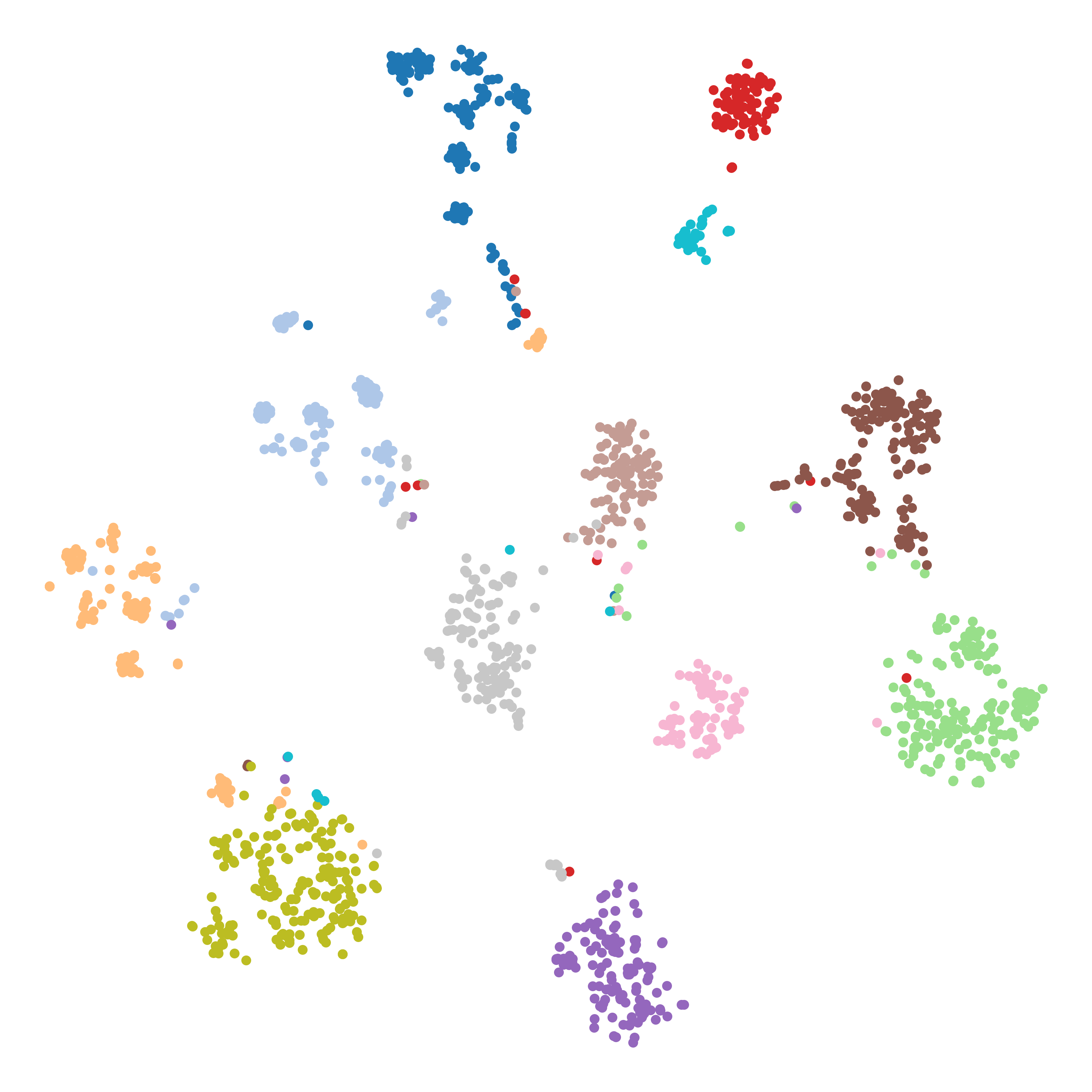}
        \caption{MCSTN -- Moderate}
        \label{fig:tsne_mcstn_moderate}
    \end{subfigure}\hfill
    \begin{subfigure}[t]{0.24\linewidth}
        \centering
        \includegraphics[width=\linewidth]{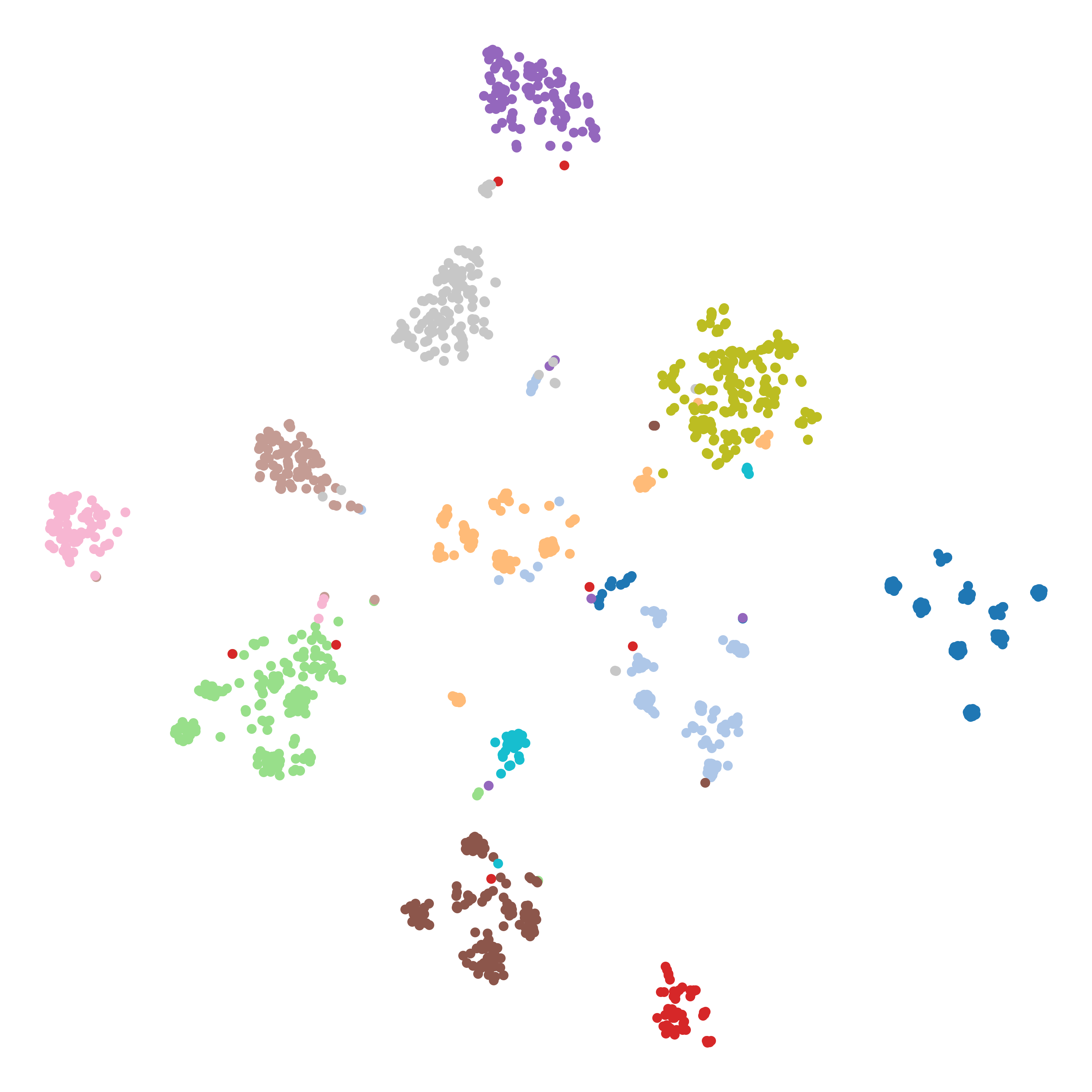}
        \caption{MCSTN -- Severe}
        \label{fig:tsne_mcstn_severe}
    \end{subfigure}

    \vspace{2mm}

    \begin{subfigure}[t]{0.24\linewidth}
        \centering
        \includegraphics[width=\linewidth]{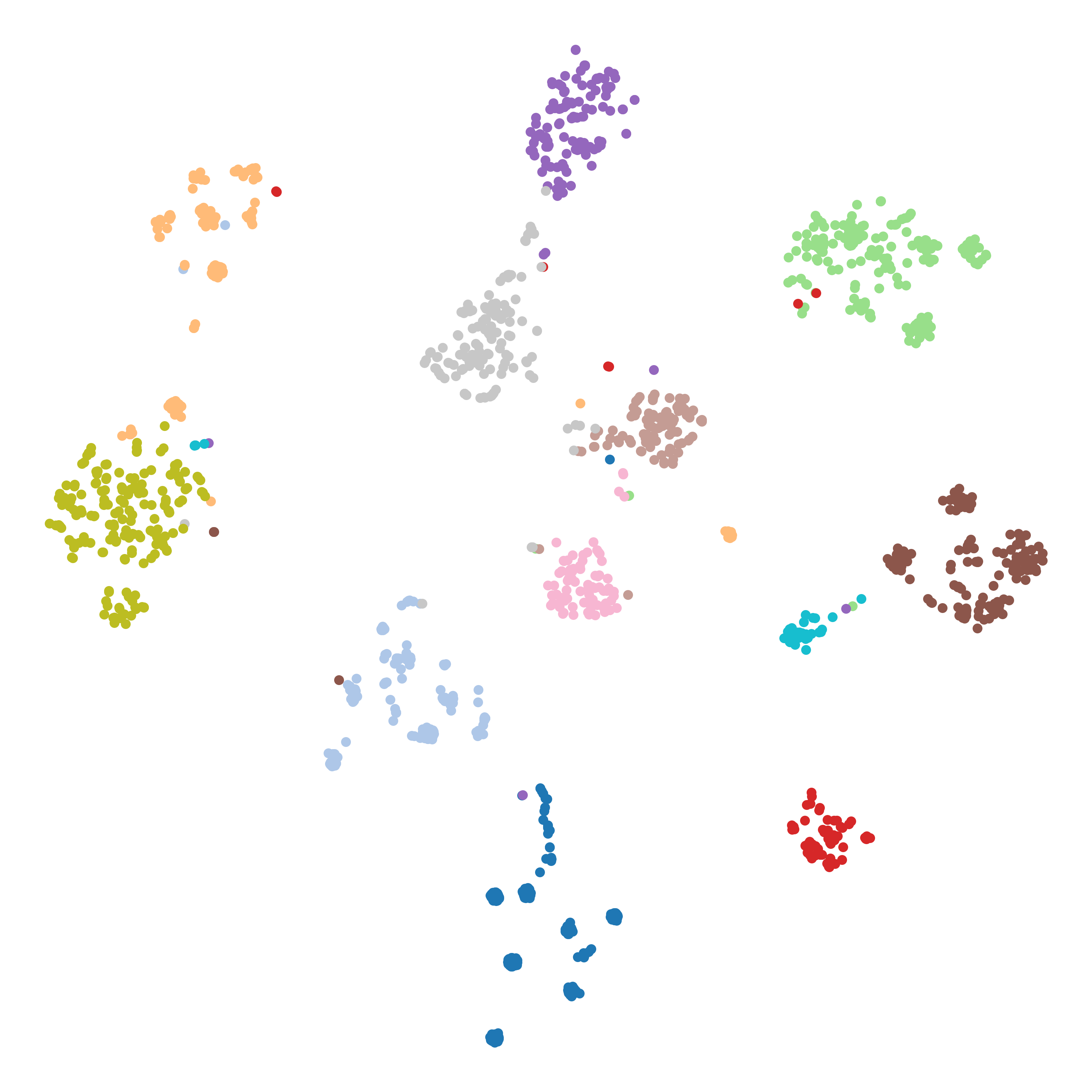}
        \caption{HARMamba -- Clean}
        \label{fig:tsne_harmamba_clean}
    \end{subfigure}\hfill
    \begin{subfigure}[t]{0.24\linewidth}
        \centering
        \includegraphics[width=\linewidth]{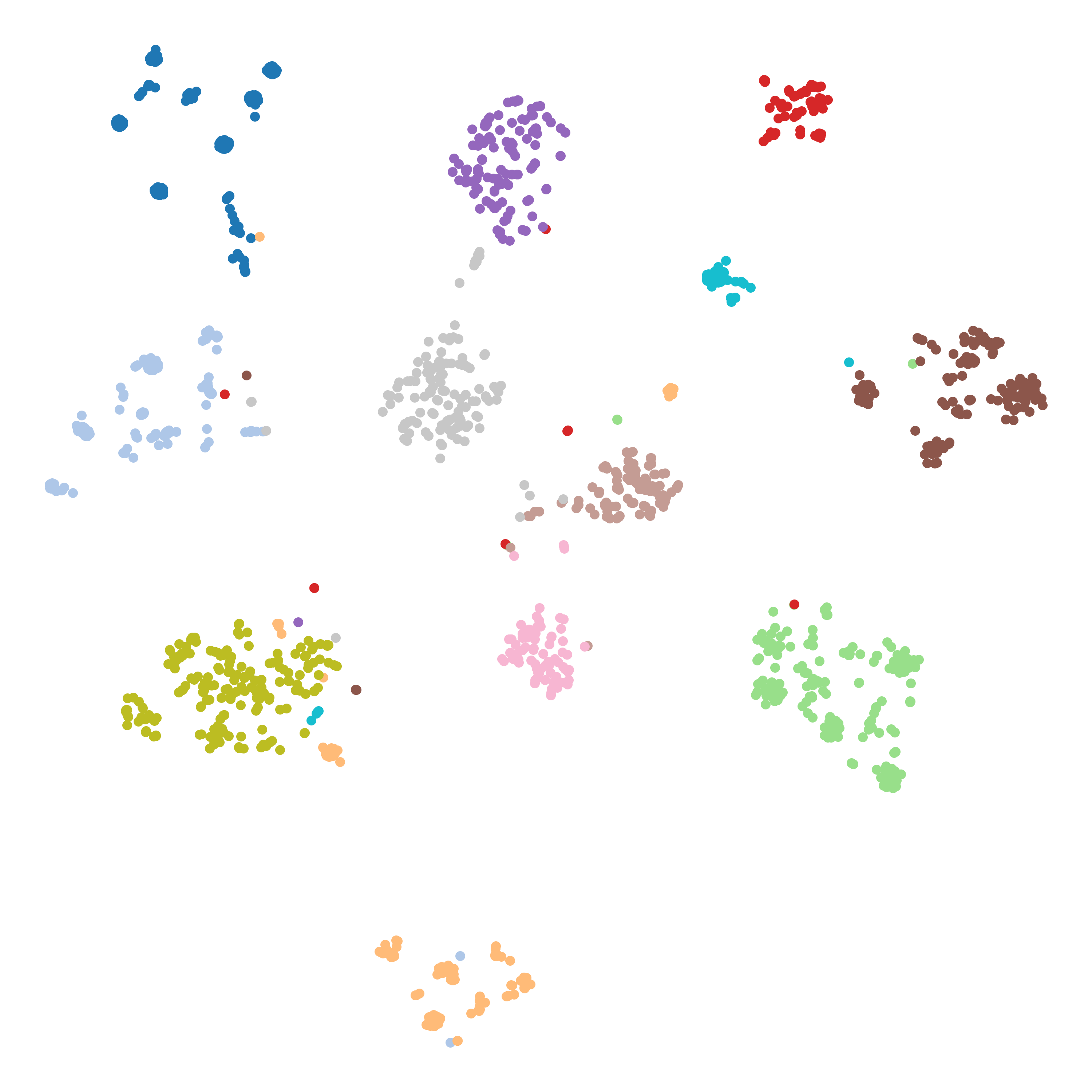}
        \caption{HARMamba -- Light}
        \label{fig:tsne_harmamba_light}
    \end{subfigure}\hfill
    \begin{subfigure}[t]{0.24\linewidth}
        \centering
        \includegraphics[width=\linewidth]{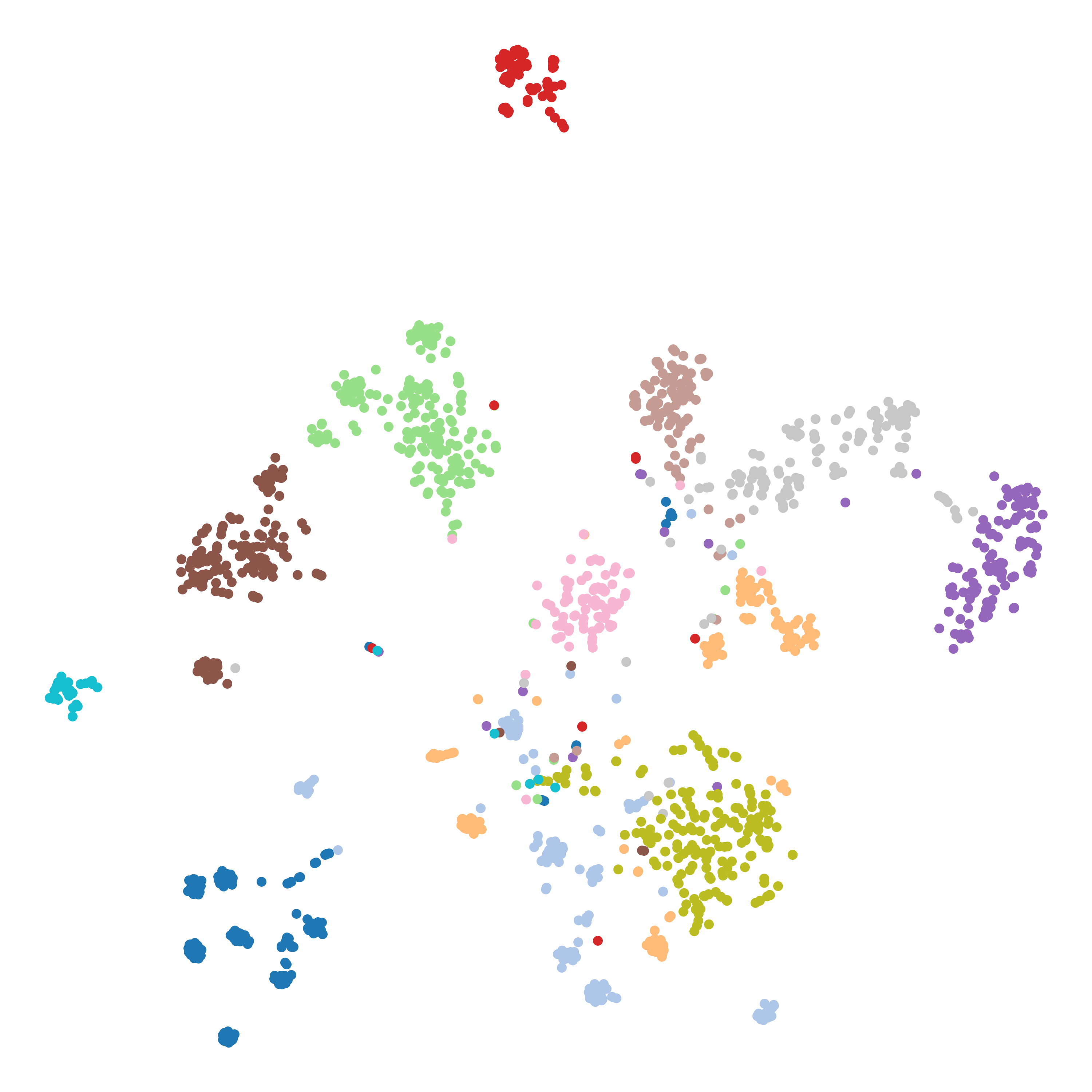}
        \caption{HARMamba -- Moderate}
        \label{fig:tsne_harmamba_moderate}
    \end{subfigure}\hfill
    \begin{subfigure}[t]{0.24\linewidth}
        \centering
        \includegraphics[width=\linewidth]{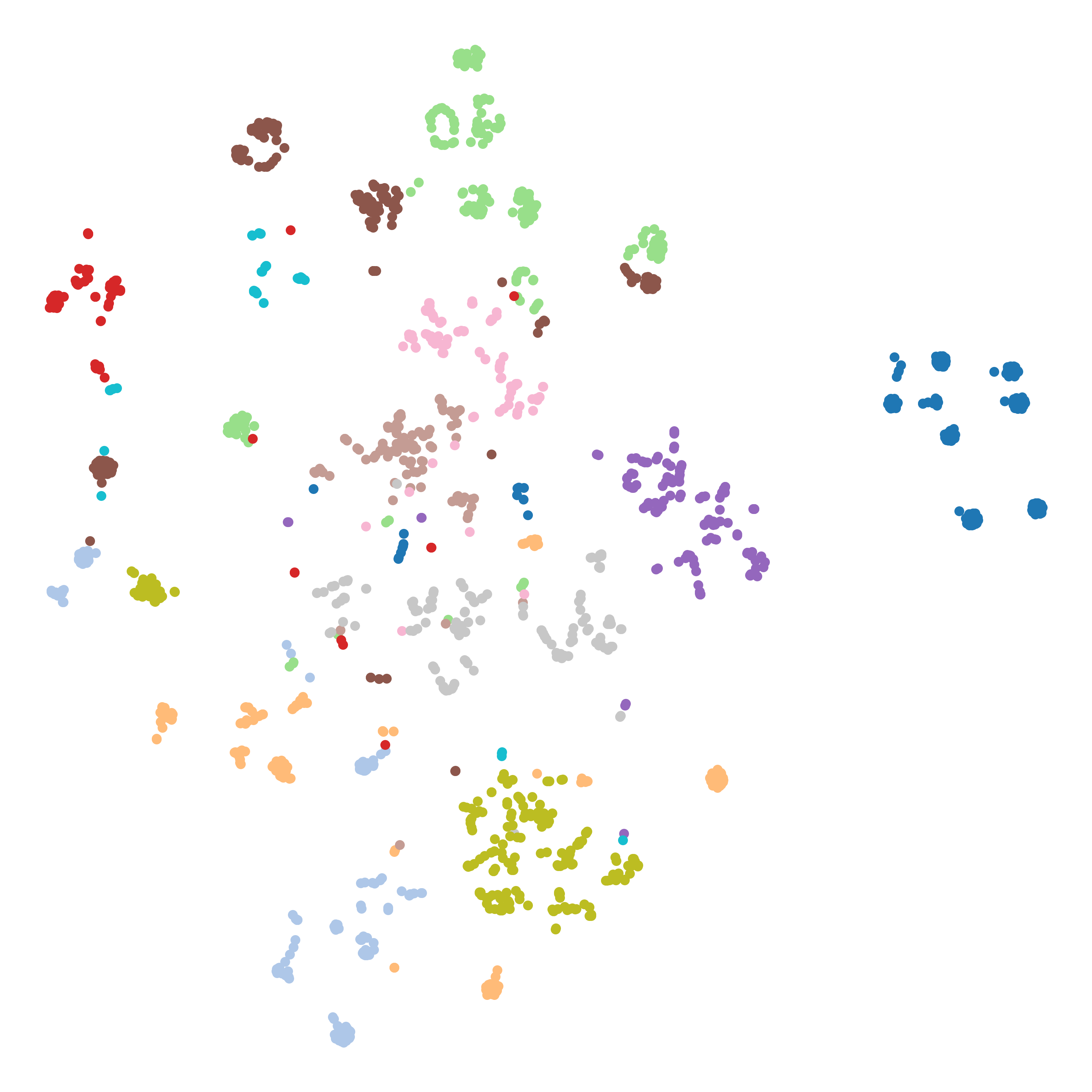}
        \caption{HARMamba -- Severe}
        \label{fig:tsne_harmamba_severe}
    \end{subfigure}

    \caption{t-SNE visualization of feature representations learned by
    MCSTN and HARMamba under clean and increasingly severe imperfect-data
    conditions on PAMAP2.}
    \label{fig:tsne_comprehensive}
\end{figure}

Fig.~\ref{fig:tsne_comprehensive} visualises the pooled representations of the PAMAP2 test set using t-SNE under increasing corruption severity. We compare MCSTN with HARMamba, which also uses a bi-directional selective state-space backbone.

Under clean input, both models form relatively well-separated class clusters. As the corruption severity increases, the MCSTN representations remain more compact and preserve clearer separation between several neighbouring classes.
The HARMamba representations show greater overlap under moderate and severe corruption. This trend is consistent with the higher Avg.\ Imp.\ and Worst scores of MCSTN on PAMAP2 in Table~\ref{tab:main_comparison}.

\subsection{Discussion and Limitations}

The imperfection generator requires the sensor topology, including the
grouping of channels into axes, modalities and nodes. These groupings are
available for the datasets used here. For less documented sensor collections,
they would need to be specified or inferred.

The synthetic corruption tests use the same five failure types as training,
restricted to the modes applicable to each dataset. The noise model uses
additive Gaussian noise controlled by a signal-to-noise schedule. Drift,
clipping, calibration errors and clock misalignment are not explicitly
modelled, so performance under these distortions remains untested.

PAMAP2 and WISDM use subject-independent evaluation. Opportunity uses a
subject-and-run split: test runs from S2 and S3 are held out, but earlier runs
from those subjects are included in training. Cross-dataset transfer is also
outside the present evaluation. These protocols therefore test different
degrees of change between the training and test data.

\section{Conclusion}
\label{conclusion}

We presented MCSTN for human activity recognition from incomplete and noisy
wearable sensor data. MCSTN conditions recognition directly on the
availability mask, combining structured imperfection modelling, a mask-aware
spatio-temporal encoder and complete-to-imperfect multi-view learning with an
EMA teacher. On PAMAP2, Opportunity and WISDM, MCSTN achieves the highest average imperfect
macro-F1 among the compared methods. On Opportunity, it also leads in
every evaluated individual failure mode, with $0.91$\,M parameters and an
inference latency of $3.78$\,ms per sample. The ablations show contributions
from the encoder branches, mask conditioning and multi-view learning, while
the cross-architecture comparison shows that structured augmentation also
improves competing models. Future work will examine drift, clipping and
clock misalignment, together with unknown sensor topologies and transfer
across datasets.


\section*{CRediT authorship contribution statement}

\textbf{Jiangtao Fan:} Conceptualization, Methodology, Validation,
Formal analysis, Investigation, Visualization, Writing -- original draft.
\textbf{Anish Jindal:} Supervision, Writing -- review \& editing.
\textbf{Amir Atapour-Abarghouei:} Supervision, Writing -- review \& editing.

\section*{Declaration of competing interest}
The authors declare that they have no known competing financial interests or personal relationships that could have appeared to influence the work reported in this paper.

\section*{Data availability}
The datasets analyzed in this study are publicly available from their respective repositories. The OPPORTUNITY, PAMAP2, and WISDM datasets can be accessed through the sources cited in the manuscript. The source code will be made publicly available upon publication.

\section*{Acknowledgements}
This work was supported by the Durham University-China Scholarship Council (CSC). The authors also acknowledge the computational resources provided by Durham University.

\bibliographystyle{elsarticle-num}
\bibliography{references}

\end{document}